\documentclass[12pt,letter,leqno]{article}
\usepackage{natbib}
\setcitestyle{authoryear,open={(},close={)},citesep={,}}
\usepackage[]{hyperref}
\usepackage{xcolor}
\hypersetup{colorlinks = true, citecolor = blue, linkcolor=blue, linkbordercolor = {white}}
\usepackage{amsmath}
\usepackage{amsfonts}
\usepackage{mathrsfs}
\usepackage{amsthm}
\usepackage{amssymb}
\usepackage{mathtools}
\usepackage{bm}
\usepackage[normalem]{ulem}
\usepackage{graphicx}
\usepackage{caption}
\usepackage{subcaption}

\usepackage
[
        a4paper,
        left=1.5cm,
        right=1.5cm,
        top=2.5cm,
        bottom=2.5cm,
]
{geometry}

\allowdisplaybreaks

\newtheorem{theorem}{Theorem}
\newtheorem{lemma}{Lemma}

\newtheorem{remark}{Remark}

\newtheorem{innercustomgeneric}{\customgenericname}
\providecommand{\customgenericname}{}
\newcommand{\newcustomtheorem}[2]{%
  \newenvironment{#1}[1]
  {%
   \renewcommand\customgenericname{#2}%
   \renewcommand\theinnercustomgeneric{##1}%
   \innercustomgeneric
  }
  {\endinnercustomgeneric}
}
\newcustomtheorem{customlemma}{Lemma}
\newcustomtheorem{customproposition}{Proposition}
\newcustomtheorem{customtheorem}{Theorem}
\newcustomtheorem{customdefinition}{Definition}
\newcustomtheorem{customexample}{Example}
\newcustomtheorem{customremark}{Remark}

\newcommand{\tran}{^{\mathstrut\scriptscriptstyle{\top}}}
\newcommand{\mX}{\mathbf{X}}
\newcommand{\mx}{\mathbf{x}}
\newcommand{\E}{{\rm E}}
\newcommand{\I}{{\rm I}}

\def\ba#1\ea{\begin{align*}#1\end{align*}}

\def\qed{\space$\square$ \par \vspace{.15in}}

\begin{document}

\Large
\centerline{\bf Hybrid deep additive neural networks}
\large
\vspace{.2cm} \centerline{Gyu Min Kim and Jeong Min Jeon}
\vspace{.2cm} \centerline{Department of Statistics, Seoul National University, South Korea}
\vspace{.2cm} \centerline{Department of Data Science, Ewha Womans University, South Korea}
\vspace{.2cm}
\small
\begin{quotation}
\noindent{\it Abstract: Traditional neural networks (multi-layer perceptrons) have become an important tool in data science due to their success across a wide range of tasks. However, their performance is sometimes unsatisfactory, and they often require a large number of parameters, primarily due to their reliance on the linear combination structure. Meanwhile, additive regression has been a popular alternative to linear regression in statistics. In this work, we introduce novel deep neural networks that incorporate the idea of additive regression. Our neural networks share architectural similarities with Kolmogorov-Arnold networks but are based on simpler yet flexible activation and basis functions. Additionally, we introduce several hybrid neural networks that combine this architecture with that of traditional neural networks. We derive their universal approximation properties and demonstrate their effectiveness through simulation studies and a real-data application. The numerical results indicate that our neural networks generally achieve much better performance than the traditional neural networks while using fewer parameters.}
\par
\vspace{.2cm}
\noindent{\it Keywords: Additive model, Basis expansion, Deep learning, Neural network}
\end{quotation}
\normalsize

\section{Introduction}

\setcounter{equation}{0}
\setcounter{subsection}{0}

Additive regression is a statistical modeling approach developed to capture nonlinear relationships between the responses and predictors without relying on strict assumptions on the form of these relationships. Instead of assuming a linear relationship, additive models allow the response variable to depend on the sum of smooth and potentially nonlinear functions of each predictor variable. There have been a number of methods to estimate additive models, such as methods based on kernel smoothing (e.g., \cite{Linton and Nielsen (1995)}, \cite{Opsomer and Ruppert (1997)}, \cite{Mammen et al. (1999)}, \cite{Jeon and Park (2020)}, \cite{Jeon et al. (2022)}) and methods based on basis expansions (e.g., \cite{Bilodeau (1992)}, \cite{Meier et al. (2009)}, \cite{Sardy and Ma (2024)}).

On the other hand, neural networks are becoming an important method in regression analysis. Its good practical performance and nice theoretical properties have been investigated by numerous works (e.g., \cite{LeCun et al. (2015)}, \cite{Schmidhuber (2015)}, \cite{Bauer and Kohler (2019)}, \cite{Schmidt-Hieber (2020)}, \cite{Kohler and Langer (2021)}). However, traditional neural networks (multi-layer perceptrons) sometimes struggle with capturing complex nonlinear relationships between predictors and responses. Additionally, they often require a huge number of parameters, which leads to high computational and memory demands. These issues mainly come from that each node does not have sufficient nonlinearity.

Recently, several modifications using nonlinear basis functions have been proposed to overcome this issue. For example, \cite{Fakhoury et al. (2022)} proposed an approach that uses a B-spline basis expansion, instead of the composition of an activation function and an affine function, to construct each node. However, this work considered only one hidden layer. \cite{Horowitz and Mammen (2007)} introduced a deep version of this neural network and suggested to use B-splines or smoothing splines to estimate their neural network. However, this work mainly focused on deriving asymptotic error rates with arbitrary estimators satisfying certain conditions. Recently, \cite{Liu et al. (2024)} proposed Kolmogorov-Arnold networks that have the same architecture as \cite{Horowitz and Mammen (2007)}. This approach uses a linear combination of the sigmoid linear unit function and a B-spline basis expansion to form each node. Although \cite{Liu et al. (2024)} demonstrates fair performance, it requires high computational cost, resulting in a long training time. Additionally, it is complicated to implement and is prone to overfitting. Moreover, its universal approximation theorem holds only for a certain class of smooth functions.

In this work, we propose an alternative approach to \cite{Liu et al. (2024)}. Specifically, 
\begin{itemize}
\item[$\bullet$] We propose a deep neural network that uses fixed activation functions and simple yet flexible basis functions, which make computation and implementation easy.
\item[$\bullet$] We introduce hybrid neural networks that combine the above neural network and the traditional neural networks, which effectively avoid overfitting.
\item[$\bullet$] We derive their universal approximation properties that hold for any continuous functions.
\end{itemize}

This paper is organized as follows. In Section \ref{methodology1}, we introduce a neural network, named as a deep additive neural network (DANN), and its property. We also introduce its hybrid variants in Section \ref{methodology2}. Section \ref{simulation} contains simulation studies and Section \ref{real data} presents real data analysis. These numerical studies show that our neural networks have better performance than the traditional neural networks in terms of both prediction errors and numbers of parameters. Section \ref{conclusion} contains conclusions, and all technical proofs are provided in the Appendix.

\section{Deep Additive Neural Networks}\label{methodology1}

Let $Y\in\mathbb{R}$ be a response and $\mathbf{X} = (X_1,\ldots,X_d)\tran\in[0,1]^d$ for $d\geq1$ be a vector of predictors. The traditional neural networks with single hidden layer assume that
\begin{eqnarray}
\E(Y|\mX) = \sum_{k=1}^p w_k\sigma\left(\sum_{j=1}^d w_{jk}X_j + b_k\right), \label{eqn:1-FCN}
\end{eqnarray}
where $p\geq1$ is the number of nodes in the hidden layer, $\sigma:\mathbb{R}\rightarrow\mathbb{R}$ is a non-polynomial continuous function, called an activation function, and $w_k, w_{jk}, b_k\in\mathbb{R}$ are parameters. The well-known universal approximation theorem (\cite{Cybenko (1989)}) tells that, for any given continuous function $f:[0,1]^d\rightarrow\mathbb{R}$ and constant $\epsilon>0$, there exist $p\geq1$ and $w_k, w_{jk}, b_k\in\mathbb{R}$ such that
\begin{align*}
\sup_{\mx=(x_1,\ldots,x_d)\tran\in[0,1]^d}\left|f(\mx) - \sum_{k=1}^p w_k\sigma\left(\sum_{j=1}^d w_{jk}x_j + b_k\right)\right| < \epsilon.
\end{align*}
In spite of this nice property, it has been known that the practical performance of (\ref{eqn:1-FCN}) is not good enough, particularly when the target regression function $\E(Y|\mX=\cdot)$ is highly nonlinear, because $\sigma$ is the only nonlinear function in (\ref{eqn:1-FCN}).

To overcome this issue, we may replace the linear functions in (\ref{eqn:1-FCN}) by nonlinear functions. Specifically, we consider the model
\begin{align}\label{eqn:O_ANN}
\E(Y|\mX) = \sum_{k=1}^p f_k\left(\sum_{j=1}^d f_{jk}(X_j)\right),
\end{align}
where $f_k:\mathbb{R}\rightarrow\mathbb{R}$ and $f_{jk}:[0,1]\rightarrow\mathbb{R}$ are possibly nonlinear functions. When $p=1$ and $f_1$ is the identity function, this model reduces to the standard additive model. By taking appropriate $f_k$ and $f_{jk}$, (\ref{eqn:O_ANN}) may approximate highly nonlinear $\E(Y|\mX=\cdot)$ better than (\ref{eqn:1-FCN}) in practice. In fact, Kolmogorov's superposition theorem (\cite{Kolmogorov (1957)}) says that any continuous function $f:[0,1]^d\rightarrow\mathbb{R}$ can be written as $\sum_{k=1}^p\phi^f_k(\sum_{j=1}^d \phi^f_{jk}(x_j))$ for some $p\geq1$ and continuous functions $\phi^f_k:\mathbb{R}\rightarrow\mathbb{R}$ and $\phi^f_{jk}:[0,1]\rightarrow\mathbb{R}$. In this work, we take $f_k$ and $f_{jk}$ that can well approximate unknown $\phi^m_k$ and $\phi^m_{jk}$, where $m:=\E(Y|\mX=\cdot)$. For this, we introduce a lemma.


\begin{figure}[!ht]
    \centering
    \begin{subfigure}[b]{0.45\textwidth}
        \includegraphics[scale = 0.9]{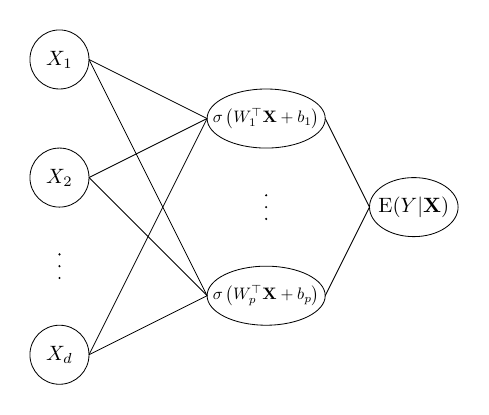}
    \end{subfigure}
    \hspace{0.1cm}
    \begin{subfigure}[b]{0.45\textwidth}
        \includegraphics[scale = 0.9]{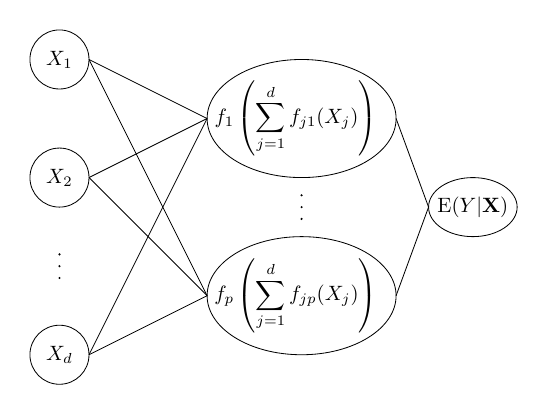}
    \end{subfigure}
    \caption{Architectures of (\ref{eqn:1-FCN}) (left) and (\ref{eqn:O_ANN}) (right). In the left panel, $W_k$ denotes $(w_{1k},\ldots,w_{dk})\tran\in\mathbb{R}^d$.}
    \label{fig:both_1}
\end{figure}

\begin{lemma}\label{lem1}
There exists a set $\{B_r:[0,1]\rightarrow\mathbb{R}\,|\,r\geq1\}$ of known functions such that, for any given continuous function $\phi:[0,1] \rightarrow \mathbb{R}$ and constant $\epsilon>0$, there exist  $q\geq1$ and $c_{r}, c \in \mathbb{R}$ such that 
\begin{align*}
\sup_{x\in[0,1]}\left|\phi(x) - \left(\sum_{r=1}^q c_{r}B_r(x) + c\right)\right| < \epsilon.
\end{align*}
\end{lemma}

\begin{figure}[!ht]
	\centering
	\includegraphics[width=\textwidth]{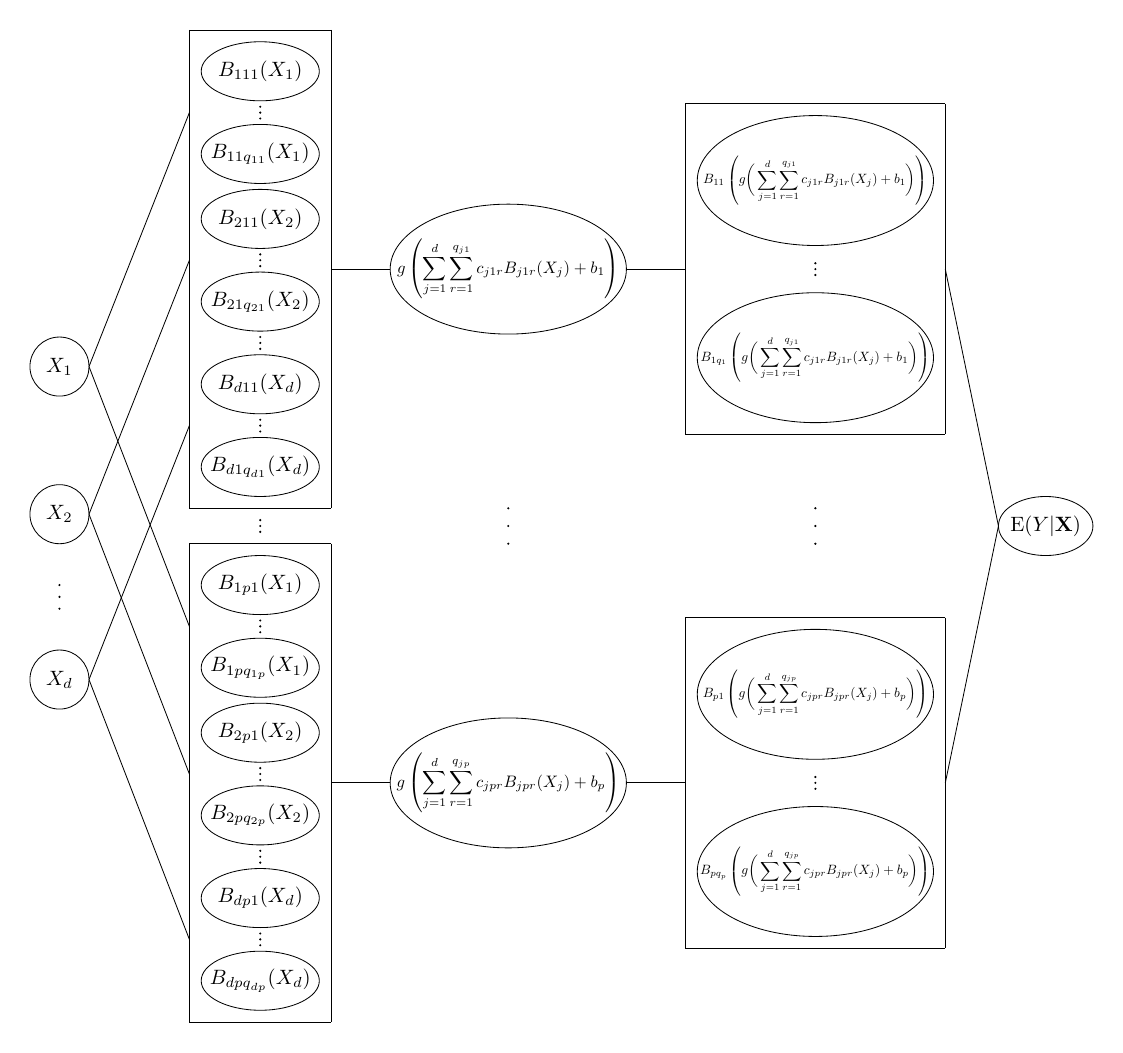}
	\caption{Architecture of ANN model.}
	\label{fig:ANN1}
\end{figure}

For example, $\{B_r(x):r\geq1\}=\{x^r:r\geq1\}$, $\{B_r(x):r\geq1\}=\{\cos(r\pi x):r\geq1\}$ and the Haar system on $[0,1]$ (e.g., \cite{Haar (1910)}), among many others, have the above property; see the proof of Lemma \ref{lem1}. In this work, we call $B_r$ satisfying the property of Lemma \ref{lem1} basis functions. From this,
\begin{align*}
\sum_{j=1}^d \phi^m_{jk}(x_j)
\end{align*}
can be well approximated by 
\begin{align*}
\sum_{j=1}^d\sum_{r=1}^{q_{jk}}c_{jkr}B_{jkr}(x_j)+b_k
\end{align*}
for some $q_{jk}\geq1$ and $c_{jkr}, b_k\in\mathbb{R}$, and any possibly different sets $\{B_{1kr}:r\geq1\},\ldots,\{B_{dkr}:r\geq1\}$ of basis functions. We also approximate $\phi^m_k:\mathbb{R}\rightarrow\mathbb{R}$ by the function
\begin{align*}
\sum_{\ell=1}^{q_k}c_{k\ell}B_{k\ell}(g(\cdot))+b,
\end{align*}
where $q_k\geq1$, $c_{k\ell},b\in\mathbb{R}$, $\{B_{k\ell}:\ell\geq1\}$ is a set of basis functions, and $g:\mathbb{R}\rightarrow[0,1]$ is a function. We call the model
\begin{align}\label{ANN}
\E(Y|\mX) = \sum_{k=1}^p\sum_{\ell=1}^{q_k}c_{k\ell}B_{k\ell}\left(g\left(\sum_{j=1}^d\sum_{r=1}^{q_{jk}}c_{jkr}B_{jkr}(X_j)+b_k\right)\right)+b
\end{align}
an Additive Neural Network (ANN). Figure \ref{fig:ANN1} illustrates the architecture of ANN. The following theorem shows that the ANN is a reasonable model.

\begin{theorem}\label{thm1}
Let $\{B_{k\ell}:\ell\geq1\}$ and $\{B_{jkr}:r\geq1\}$ be any sets of basis functions and $g$ be non-constant and Lipschitz continuous. Then, for any given continuous function $f:[0,1]^d \rightarrow \mathbb{R}$ and constant $\epsilon>0$, there exist $p, q_k, q_{jk}\geq1$ and $c_{k\ell}, c_{jkr}, b_k, b\in\mathbb{R}$ such that
\begin{align*}
\sup_{\mx\in[0,1]^d}\left|f(\mx) - \left(\sum_{k=1}^p\sum_{\ell=1}^{q_k}c_{k\ell}B_{k\ell}\left(g\left(\sum_{j=1}^d\sum_{r=1}^{q_{jk}}c_{jkr}B_{jkr}(x_j)+b_k\right)\right)+b\right)\right| < \epsilon.
\end{align*}
\end{theorem}

Examples of $g$ satisfying the assumption in Theorem \ref{thm1} include the logistic function, $(\rm{tanh}+1)/2$ and any Lipschitz continuous cumulative distribution function, where tanh is the hyperbolic tangent function. The hyperparameters of ANN that we need to choose include $p$, $q_k$, $q_{jk}$, $g$ and the sets of basis functions. For given hyperparameters, we estimate $c_{k\ell}$, $c_{jkr}$, $b_k$ and $b$ in (\ref{ANN}) by finding values that minimize $\sum_{i=1}^n(Y_i-\hat{Y}_i)^2$. For this, we initialize them and then update subsequently using an optimization algorithm.

\begin{figure}[!ht]
	\centering
	\includegraphics[width=\textwidth]{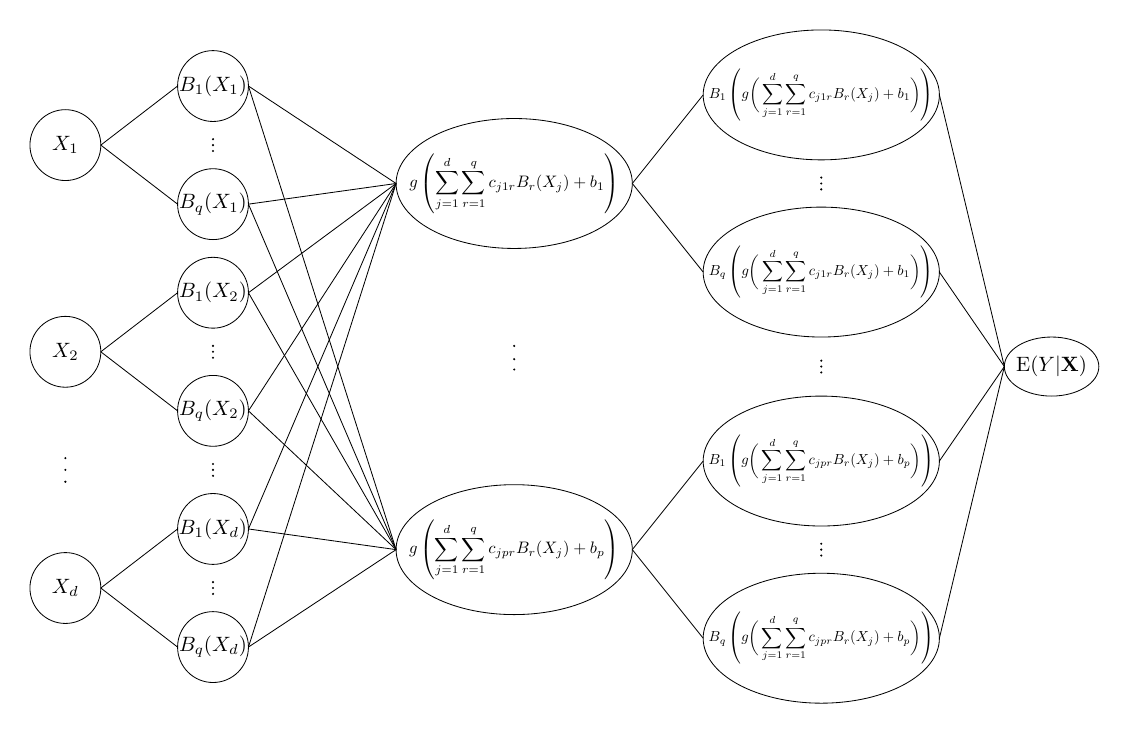}
	\caption{Architecture of ANN model in the case of the same set and same number of basis functions.}
	\label{fig:ANN2}
\end{figure}

\begin{remark}\label{rmk1}
The ANN has important advantages.
\begin{itemize}
\item[1.] First, the ANN architecture is easy to build. Note that $\{B_{jkr}(X_j):1\leq j\leq d, 1\leq r\leq q_{jk}\}$ for $1\leq k\leq p$ that are in the left rectangles of Figure \ref{fig:ANN1} can be easily obtained from $\mX$. Similarly, $\{B_{k\ell}(g(\sum_{j=1}^d\sum_{r=1}^{q_{jk}}c_{jkr}B_{jkr}(X_j)+b_k)):1\leq\ell\leq q_k\}$ that is in the $k$th right rectangle of Figure \ref{fig:ANN1} can be easily obtained from $g(\sum_{j=1}^d\sum_{r=1}^{q_{jk}}c_{jkr}B_{jkr}(X_j)+b_k)$ for $1\leq k\leq p$. If we use the same basis set, say $\{B_r:r\geq1\}$, and the same number of basis functions, say $q$, across all nodes, then the ANN architecture is even more simplified; see Figure \ref{fig:ANN2}.
\item[2.] Second, we can directly apply optimization algorithms used for the traditional neural networks, such as ADAM (\cite{Kingma and Ba (2017)}), to update the parameters of ANN. This is because $\sum_{j=1}^d\sum_{r=1}^{q_{jk}}c_{jkr}B_{jkr}(X_j)+b_k$ is simply the linear combination of $\{B_{jkr}(X_j):1\leq j\leq d, 1\leq r\leq q_{jk}\}\cup\{1\}$ for $1\leq k\leq p$, and
\begin{align*}
\sum_{k=1}^p\sum_{\ell=1}^{q_k}c_{k\ell}B_{k\ell}\left(g\left(\sum_{j=1}^d\sum_{r=1}^{q_{jk}}c_{jkr}B_{jkr}(X_j)+b_k\right)\right)+b
\end{align*}
is simply the linear combination of $\{B_{k\ell}(g(\sum_{j=1}^d\sum_{r=1}^{q_{jk}}c_{jkr}B_{jkr}(X_j)+b_k)):1\leq k\leq p, 1\leq\ell\leq q_k\}\cup\{1\}$.
\item[3.] Cosine basis might approximate a periodic function well and the Haar basis might approximate a non-smooth function well. By mixing various types of basis, the ANN might approximate diverse functions well in practice.
\end{itemize}
\end{remark}

\begin{figure}[!ht]
	\centering
	\includegraphics[width=\textwidth]{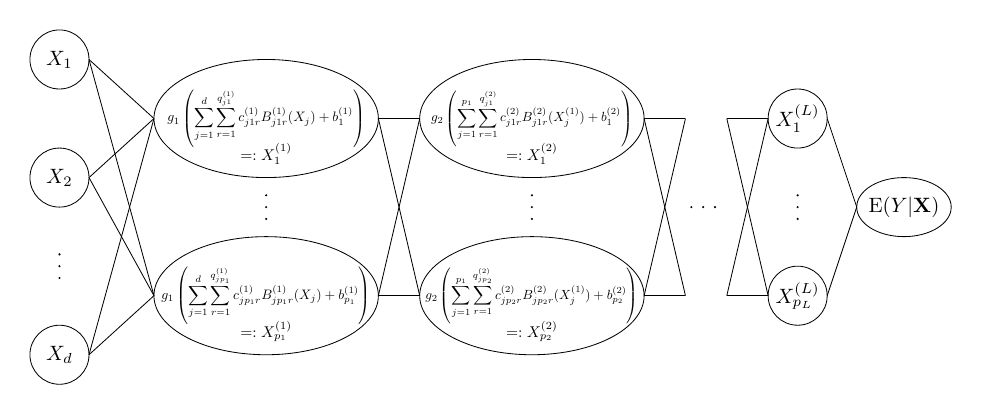}
	\caption{Architecture of DANN model.}
	\label{fig:ADNN}
\end{figure}

It has been known that the traditional neural networks with multiple hidden layers, sometimes called a Deep Neural Network (DNN), tends to provide better performance than that with single hidden layer. This motivates us to add more hidden layers to the ANN. The ANN with $L$ hidden layers is illustrated in Figure \ref{fig:ADNN}. Formally, it can be written as
\begin{align}\label{eqn:ADNN}
\begin{split}
\E(Y|\mX) & = \sum_{k=1}^{p_L}\sum_{r=1}^{q_k}c_{kr}B_{kr}(X_{k}^{(L)})+b, \\
X_k^{(L)} & = g_L\left(\sum_{j=1}^{p_{L-1}}\sum_{r=1}^{q_{jk}^{(L)}}c_{jkr}^{(L)}B_{jkr}^{(L)}\left(X_j^{(L-1)}\right)+b_k^{(L)}\right) \text{ for } k = 1, \ldots, p_L, \\
& \vdots \\
X_k^{(1)} & = g_1\left(\sum_{j=1}^{d}\sum_{r=1}^{q_{jk}^{(1)}}c_{jkr}^{(1)}B_{jkr}^{(1)}(X_j)+b_k^{(1)}\right) \text{ for } k = 1, \ldots, p_1,
\end{split}
\end{align}
where $p_l$ is the number of nodes in the $l$th hidden layer, $\{B_{kr}:r\geq1\}$ and $\{B_{jkr}^{(l)}:r\geq1\}$ are sets of basis functions, $q_k$ and $q_{jk}^{(l)}$ are the numbers of used basis functions, $g_l:\mathbb{R}\rightarrow[0,1]$ are functions, and $c_{kr}, c_{jkr}^{(l)}, b_k^{(l)}, b\in\mathbb{R}$ are parameters. We call (\ref{eqn:ADNN}) a Deep Additive Neural Network (DANN).

The advantages of ANN given in Remark \ref{rmk1} are still valid for the DANN. In particular, constructing basis functions in this network is easier than that in \cite{Liu et al. (2024)} since the output values from $g_l$ lie in $[0,1]$, and hence we do not need to adapt basis functions in the next layer according to the output values. In \cite{Liu et al. (2024)}, however, knots for the B-spline basis need to be adjusted according to the output values from their activation functions. 

\section{Hybrid Networks}\label{methodology2}

The DANN may approximate a complex regression function well, but its complexity might be too high for a relatively simple regression function. In the latter case, it can overfit data. To adjust the model complexity, we introduce three networks that combine the architectures of ANN and DNN. These networks use the architecture of ANN for some layers and use that of DNN for other layers.

\begin{figure}[!ht]
	\centering
	\includegraphics[width=\textwidth]{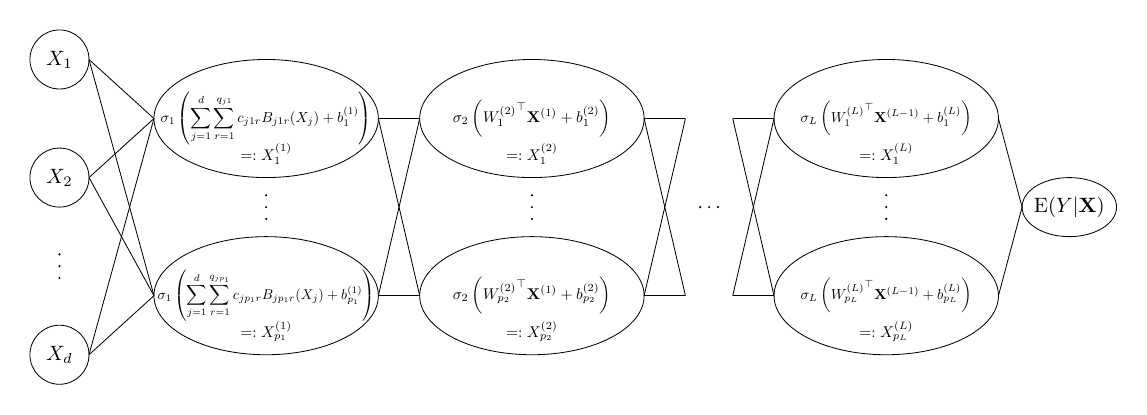}
	\caption{Architecture of HDANN1 model.}
	\label{fig:HADNN1}
\end{figure}

The first hybrid network, namely the Hybrid Deep Additive Neural Network 1 (HDANN1), uses the ANN architecture to construct the first hidden layer and uses the DNN architecture for the remaining layers. This network is illustrated in Figure \ref{fig:HADNN1}. Formally, it can be written as
\begin{align}\label{eqn:HADNN1}
\begin{split}
\E(Y|\mX) & = \sum_{k=1}^{p_L}w_k X_{k}^{(L)}+b, \\
X_k^{(L)} & = \sigma_L\left({W_k^{(L)}}\tran\mathbf{X}^{(L-1)} + b_k^{(L)}\right)  \text{ for } k = 1, \ldots, p_L, \\
& \vdots \\
X_k^{(2)} & = \sigma_2\left({W_k^{(2)}}\tran\mathbf{X}^{(1)}+b_k^{(2)}\right) \text{ for } k = 1, \ldots, p_2, \\
X_k^{(1)} & = \sigma_1\left(\sum_{j=1}^{d}\sum_{r=1}^{q_{jk}}c_{jkr}B_{jkr}(X_j)+b_k^{(1)}\right) \text{ for } k = 1, \ldots, p_1,
\end{split}
\end{align}
where $W_k^{(l)} \in \mathbb{R}^{p_{l-1}}$ are vectors of weight parameters, $\mathbf{X}^{(l)}=\big(X_1^{(l)},\ldots,X_{p_l}^{(l)}\big)\tran\in\mathbb{R}^{p_l}$, and $\sigma_l:\mathbb{R}\rightarrow\mathbb{R}$ are functions.

This network also has the universal approximation property even when $p_1=\cdots=p_L$ and $\sigma_1=\cdots=\sigma_L$. To describe this, let $\mathbf{q}$ denote the collection of $q_{jk}$, $\mathbf{W}$ denote the collection of $W_k^{(l)}$, $\mathbf{c}$ denote the collection of $c_{jkr}$, and $\mathbf{b}$ denote the collection of $b_k^{(l)}$ and $b$. Also, let
\begin{align*}
{\rm HDANN}1_{L,\mathbf{q},\mathbf{W},\mathbf{c},\mathbf{b}}(\mx)=\sum_{k=1}^{p_L}w_k x_{k}^{(L)}+b
\end{align*}
denote the function on $[0,1]^d$ defined through the architecture at (\ref{eqn:HADNN1}).

\begin{theorem}\label{thm2}
Let $\{B_{jkr}:r\geq1\}$ be any sets of basis functions, $p_1=\cdots=p_L\geq d+3$, and $\sigma_1=\cdots=\sigma_L$ be a non-affine Lipschitz continuous function which is continuously differentiable at at least one point, with nonzero derivative at that point. Then, for any given continuous function $f:[0,1]^d \rightarrow \mathbb{R}$ and constant $\epsilon>0$, there exist $L, q_{jk}\geq1$, $W_k^{(l)}\in \mathbb{R}^{p_{l-1}}$ and $c_{jkr}, b_k^{(l)}, b\in\mathbb{R}$ such that
\begin{align*}
\sup_{\mx\in[0,1]^d}\left|f(\mx) - {\rm HDANN}1_{L,\mathbf{q},\mathbf{W},\mathbf{c},\mathbf{b}}(\mx)\right| < \epsilon.
\end{align*}
\end{theorem}

Examples of $\sigma_l$ satisfying the assumption in Theorem \ref{thm2} include the logistic function, the rectified linear unit (ReLU) function and tanh.

\begin{figure}[!ht]
	\centering
	\includegraphics[width=\textwidth]{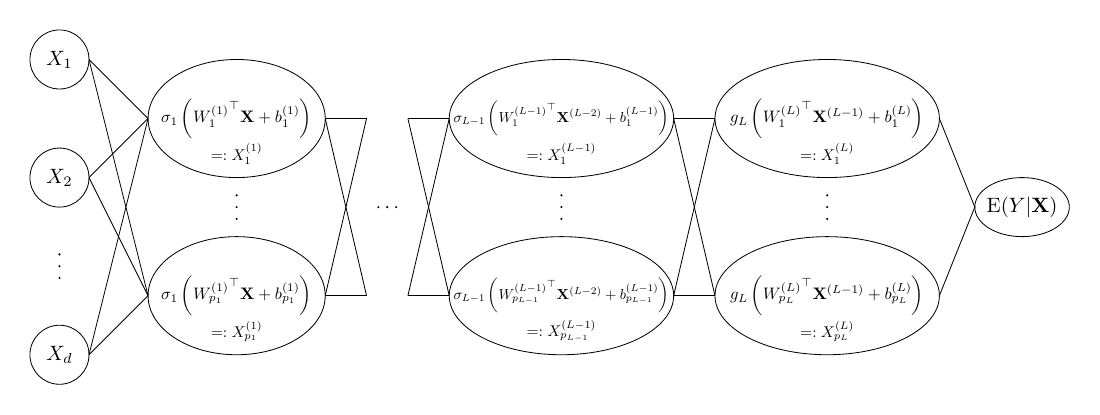}
	\caption{Architecture of HDANN2 model.}
	\label{fig:HADNN2}
\end{figure}

The second hybrid network, namely the Hybrid Deep Additive Neural Network 2 (HDANN2), is the reverse version of HDANN1. It adopts the ANN architecture to construct the output layer and the DNN architecture for the hidden layers. This network is written below and visualized in Figure \ref{fig:HADNN2}.
\begin{align}\label{eqn:HADNN2}
\begin{split}
\E(Y|\mX) & = \sum_{k=1}^{p_L}\sum_{r=1}^{q_k}c_{kr}B_{kr}(X_{k}^{(L)})+b, \\
X_k^{(L)} & = g_L\left({W_k^{(L)}}\tran\mathbf{X}^{(L-1)}+b_k^{(L)}\right)  \text{ for } k = 1, \ldots, p_L, \\
X_k^{(L-1)} & = \sigma_{L-1}\left({W_k^{(L-1)}}\tran\mathbf{X}^{(L-2)}+b_k^{(L-1)}\right) \text{ for } k = 1, \ldots, p_{L-1}, \\
& \vdots  \\
X_k^{(1)} & = \sigma_1\left({W_k^{(1)}}\tran\mathbf{X}+b_k^{(1)}\right) \text{ for } k = 1, \ldots, p_1,
\end{split}
\end{align}
where $g_L:\mathbb{R}\rightarrow[0,1]$ and $\sigma_l:\mathbb{R}\rightarrow\mathbb{R}$ are functions. 

This network also has the universal approximation property even when $p_1=\cdots=p_L$ and $\sigma_1=\cdots=\sigma_{L-1}=g_L$. Define $\mathbf{q},\mathbf{W},\mathbf{c}$ and $\mathbf{b}$ similarly as in the case of HDANN1 and let
\begin{align*}
{\rm HDANN}2_{L,\mathbf{q},\mathbf{W},\mathbf{c},\mathbf{b}}(\mx)=\sum_{k=1}^{p_L}\sum_{r=1}^{q_k}c_{kr}B_{kr}(x_{k}^{(L)})+b
\end{align*}
denote the function on $[0,1]^d$ defined through the architecture at (\ref{eqn:HADNN2}).

\begin{theorem}\label{thm3}
Let $\{B_{kr}:r\geq1\}$ be any sets of basis functions, $p_1=\cdots=p_L\geq d+3$, and $\sigma_1=\cdots=\sigma_{L-1}=g_L$ be a continuous function which is continuously differentiable at at least one point, with nonzero derivative at that point. Then, for any given continuous function $f:[0,1]^d \rightarrow \mathbb{R}$ and constant $\epsilon>0$, there exist $L, q_k\geq1$, $W_k^{(l)}\in \mathbb{R}^{p_{l-1}}$ and $c_{kr}, b_k^{(l)}, b\in\mathbb{R}$ such that
\begin{align*}
\sup_{\mx\in[0,1]^d}\left|f(\mx) - {\rm HDANN}2_{L,\mathbf{q},\mathbf{W},\mathbf{c},\mathbf{b}}(\mx)\right| < \epsilon.
\end{align*}
\end{theorem}

\begin{figure}[!ht]
	\centering
	\includegraphics[width=\textwidth]{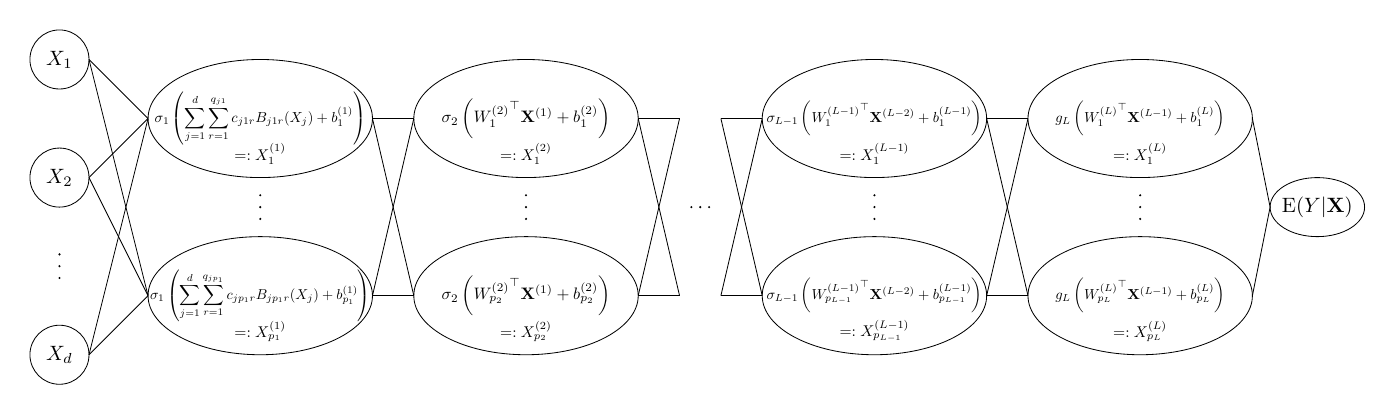}
	\caption{Architecture of HDANN3 model.}
	\label{fig:HADNN3}
\end{figure}

The third hybrid network, namely the Hybrid Deep Additive Neural Network 3 (HDANN3), combines the HDANN1 and HDANN2. This network takes the ANN architecture for the first hidden layer and output layer and takes the DNN architecture for the remaining layers. This network is formulated below and depicted in Figure \ref{fig:HADNN3}.
\begin{align}\label{eqn:HADNN3}
\begin{split}
\E(Y|\mX) & = \sum_{k=1}^{p_L}\sum_{r=1}^{q_k}c_{kr}B_{kr}(X_{k}^{(L)})+b, \\
X_k^{(L)} & = g_L\left({W_k^{(L)}}\tran\mathbf{X}^{(L-1)}+b_k^{(L)}\right)  \text{ for } k = 1, \ldots, p_L, \\
X_k^{(L-1)} & = \sigma_{L-1}\left({W_k^{(L-1)}}\tran\mathbf{X}^{(L-2)}+b_k^{(L-1)}\right) \text{ for } k = 1, \ldots, p_{L-1}, \\
& \vdots \\
X_k^{(2)} & = \sigma_2\left({W_k^{(2)}}\tran\mathbf{X}^{(1)}+b_k^{(2)}\right) \text{ for } k = 1, \ldots, p_2, \\
X_k^{(1)} & = \sigma_1\left(\sum_{j=1}^{d}\sum_{r=1}^{q_{jk}}c_{jkr}B_{jkr}(X_j)+b_k^{(1)}\right) \text{ for } k = 1, \ldots, p_1.
\end{split}
\end{align}

Note that the HDANN3 reduces to the ANN, which already has the universal approximation property, when there is a single hidden layer ($L=1$).



\section{Simulations}\label{simulation}

In this section, we compare the proposed networks (DANN, HDANN1, HDANN2 and HDANN3) with the DNN. We compare not only their prediction performance but also the numbers of parameters. We generated $Y$ from the following two models:
\begin{align}\label{data model}
\begin{split}
\text{Model 1: }Y &= \exp\left(\sum_{j=1}^3 X_j^3-\sum_{j=4}^6 X_j^3\right) + \varepsilon,\\
\text{Model 2: }Y &= \left(1 + X_1 + 2X_2^2 + 3X_3^3 - \exp(X_4) - \log(X_5+1) - \left\vert X_6-0.5\right\vert\right)^2 + \varepsilon,
\end{split}
\end{align}
where $X_j$ were generated from the uniform distribution $U(0,1)$ and the error term $\varepsilon$ was generated from the normal distribution $N(0,0.1^2)$ independently of $\mX$.

To reduce the number of hyperparameter combinations, we used the same number of nodes and the same activation function across all hidden layers, that is, $p_l\overset{l}{\equiv}p$ and $\sigma_l\overset{l}{\equiv}\sigma$. For the DNN, we explored all combinations of $L \in \{2\times t:1\leq t\leq 9\}$, $p \in \{2^t:t=3,5,7,9,11\}$ and $\sigma \in\{ \text{logistic, ReLU, tanh}\}$. This resulted in $9\times5\times3=135$ combinations for the DNN. For the proposed networks, we used the logistic function for all $g_l$. We also used the same set $\{B_r:r\geq1\}$ of basis functions and the same number $q$ of basis functions across all layers and nodes. We investigated all combinations of $L \in \{1,3,5,7,9\}$, $p \in \{2^t:t=2,4,6,8,10\}$, $q \in \{3,5,7,9,11\}$ and $\sigma \in\{ \text{logistic, ReLU, tanh}\}$, along with two basis options, $\{B_r(x):r\geq1\}=\{x^r:r\geq1\}$ (polynomial) and $\{B_r(x):r\geq1\}=\{\cos(r\pi x):r\geq1\}$ (cosine). This resulted in $5\times 5\times 5\times 3\times 2=750$ combinations for each proposed network.

In this setting, the number of parameters for each network is given by
\begin{align*}
\#\text{DNN}(d,L,p)&=(d+1)p+(p+1)p(L-1)+p+1,\\
\#\text{DANN}(d,L,p,q)&=(dq+1)p+(pq+1)p(L-1)+pq+1,\\
\#\text{HDANN1}(d,L,p,q)&=(dq+1)p+(p+1)p(L-1)+p+1,\\
\#\text{HDANN2}(d,L,p,q)&=(d+1)p+(p+1)p(L-1)+pq+1,\\
\#\text{HDANN3}(d,L,p,q)&=(dq+1)p+(p+1)p(L-1)+pq+1.
\end{align*}
With the largest $L, p$ and $q$ for each network, it holds that
\begin{align*}
&\#\text{DANN}(6,9,2^{10},11)>\#\text{DNN}(6,18,2^{11})\\
&>\#\text{HDANN3}(6,9,2^{10},11)>\#\text{HDANN1}(6,9,2^{10},11)>\#\text{HDANN2}(6,9,2^{10},11).
\end{align*}

For each model in (\ref{data model}), we generated 5 Monte-Carlo samples, with each sample consisting of a training set, a validation set and a test set. The sizes of the validation sets and test sets were fixed at 500, while the sizes of the training sets were either 1000 (scenario 1) or 2000 (scenario 2). In each training process, we used standardized response values $Y^{\rm st}_i$. Here, the mean and standard deviation for the standardization were computed from the training set. The parameters were initialized using the Xavier uniform initialization (\cite{Glorot and Bengio (2010)}) and then updated in the direction of minimizing $\sum_{i\in I_{\rm train}}(Y^{\rm st}_i-\hat{Y}^{\rm st}_i)^2$ using the ADAM optimization with a batch size of 512 and a learning rate of $10^{-4}$, where $I_{{\rm train}}$ is the index set of the training set. To reduce computation time, we terminated the updating process if $\sum_{i\in I_{\rm train}}(Y^{\rm st}_i-\hat{Y}^{\rm st}_i)^2/|I_{\rm train}|$ does not decrease by more than $10^{-3}$ over the recent 10 updates, where $|I_{{\rm train}}|$ denotes the cardinality of $I_{{\rm train}}$.

\begin{figure}[!ht]
    \centering
    \includegraphics[width=0.475\textwidth]{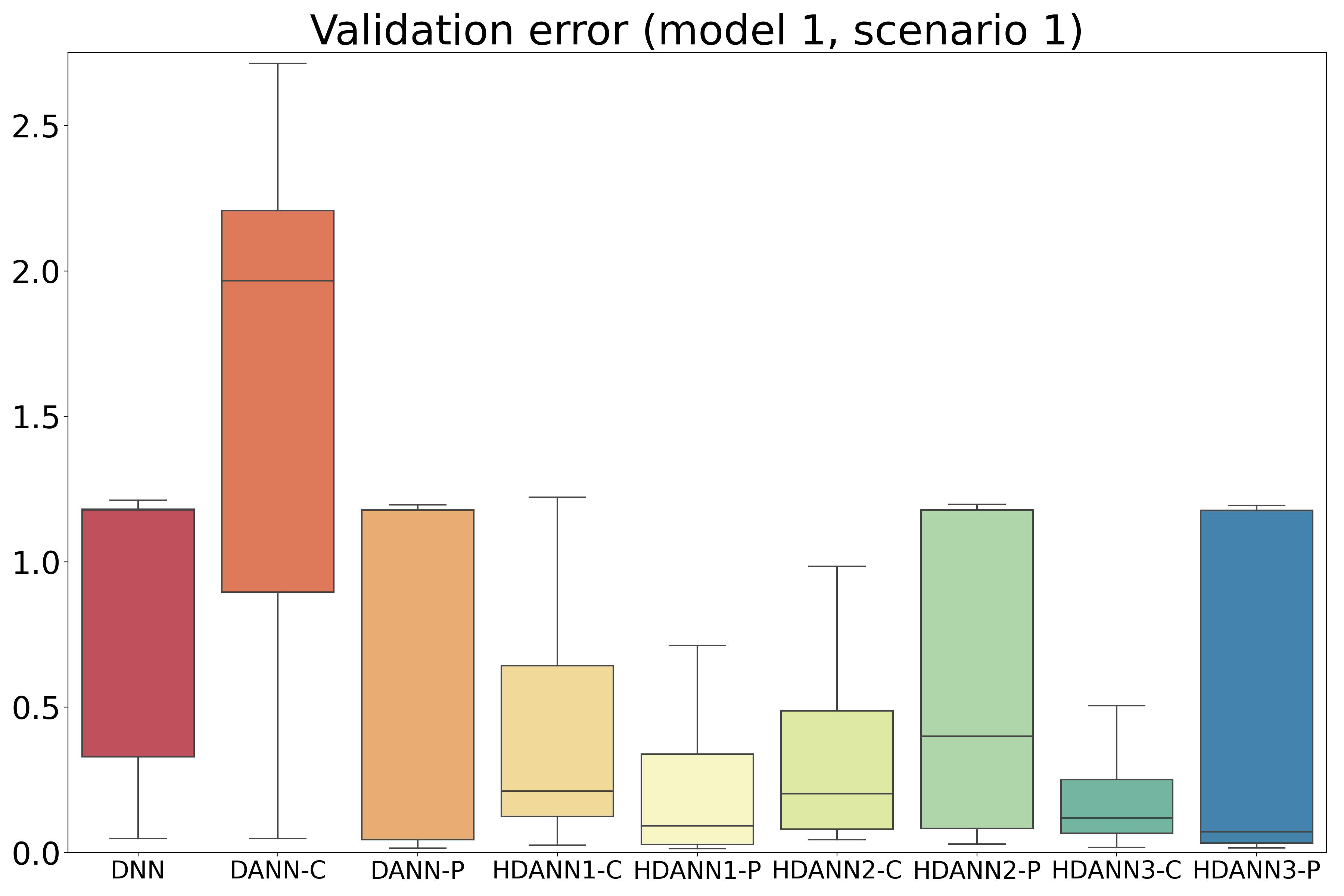}
    \includegraphics[width=0.475\textwidth]{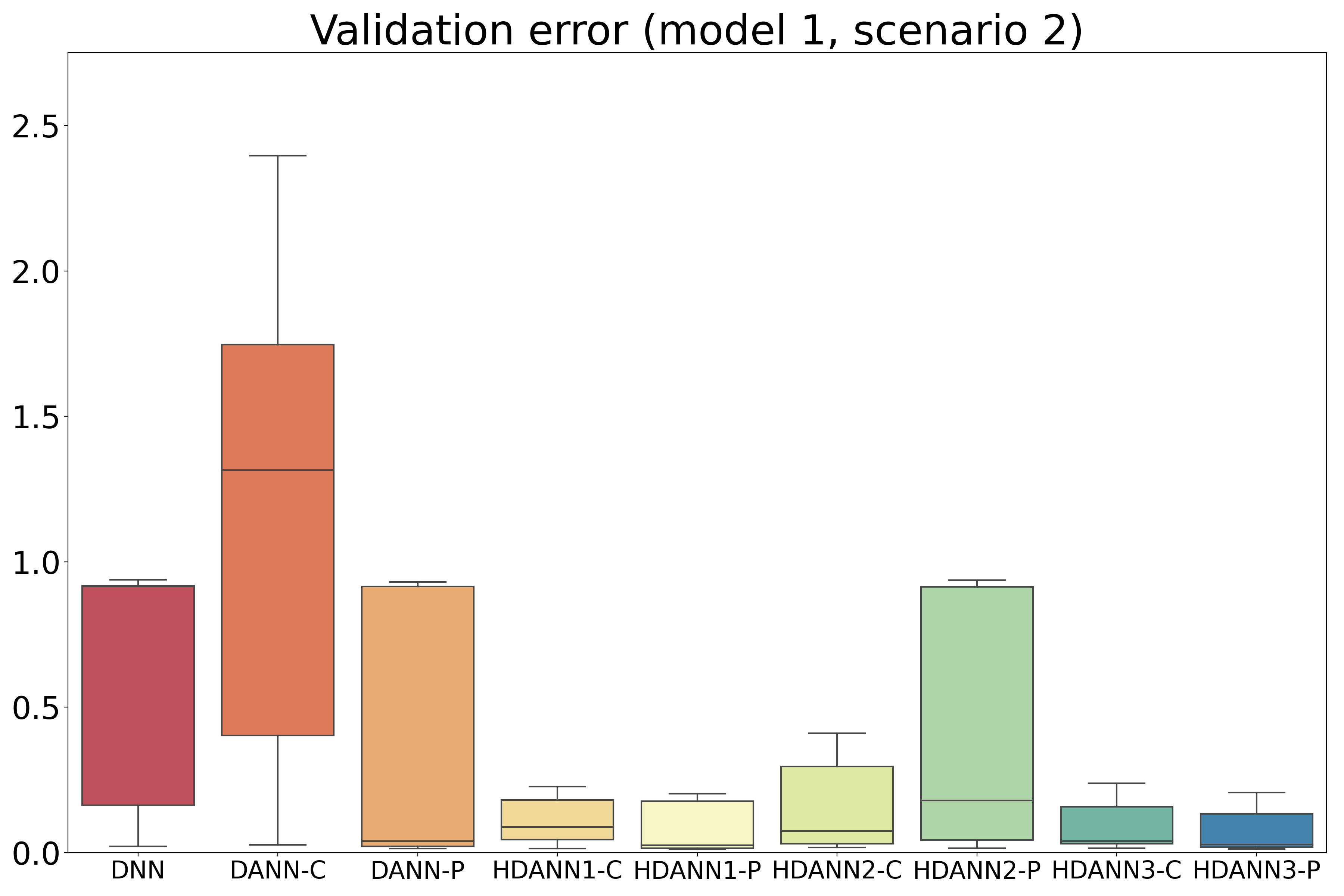}
    \includegraphics[width=0.475\textwidth]{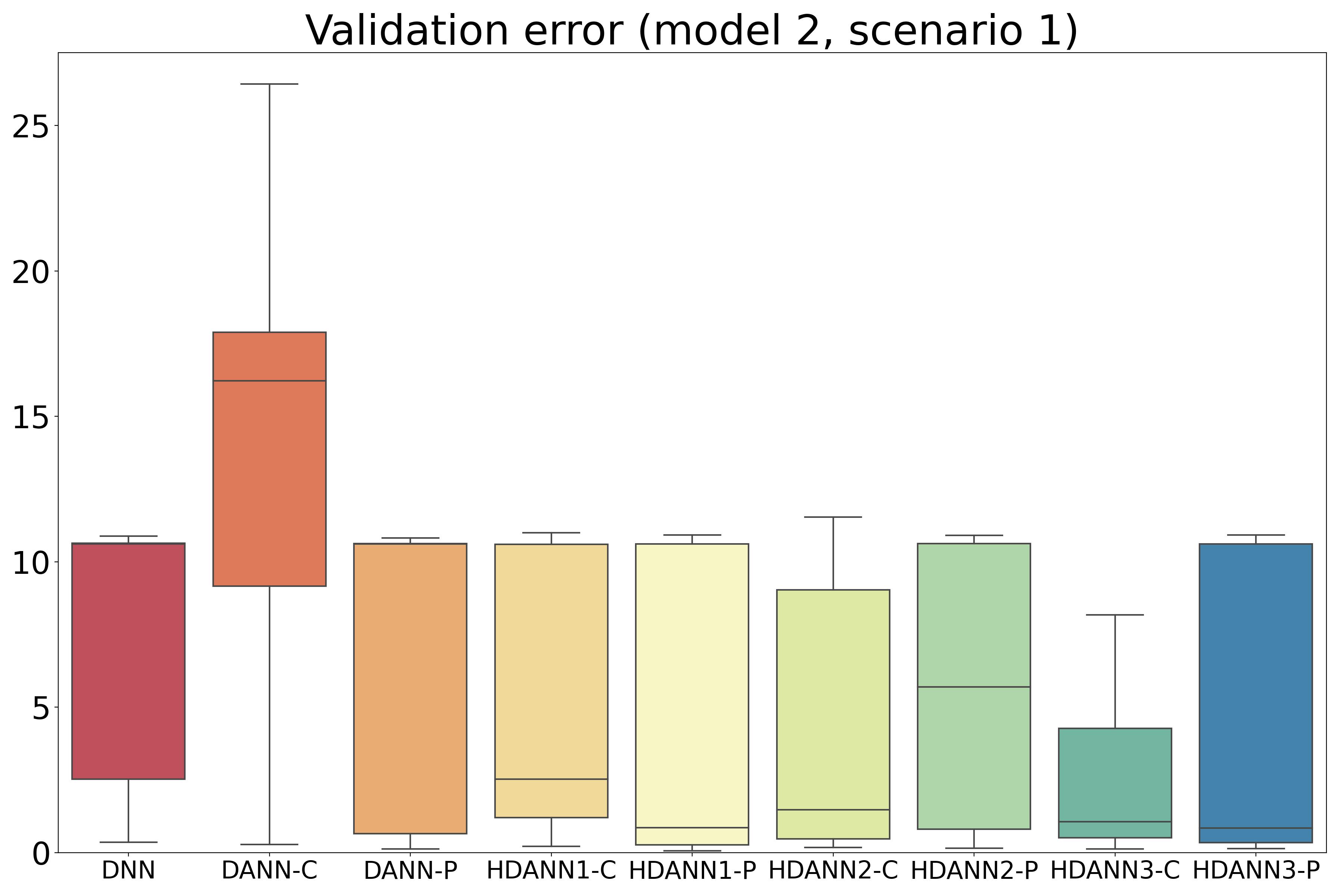}
    \includegraphics[width=0.475\textwidth]{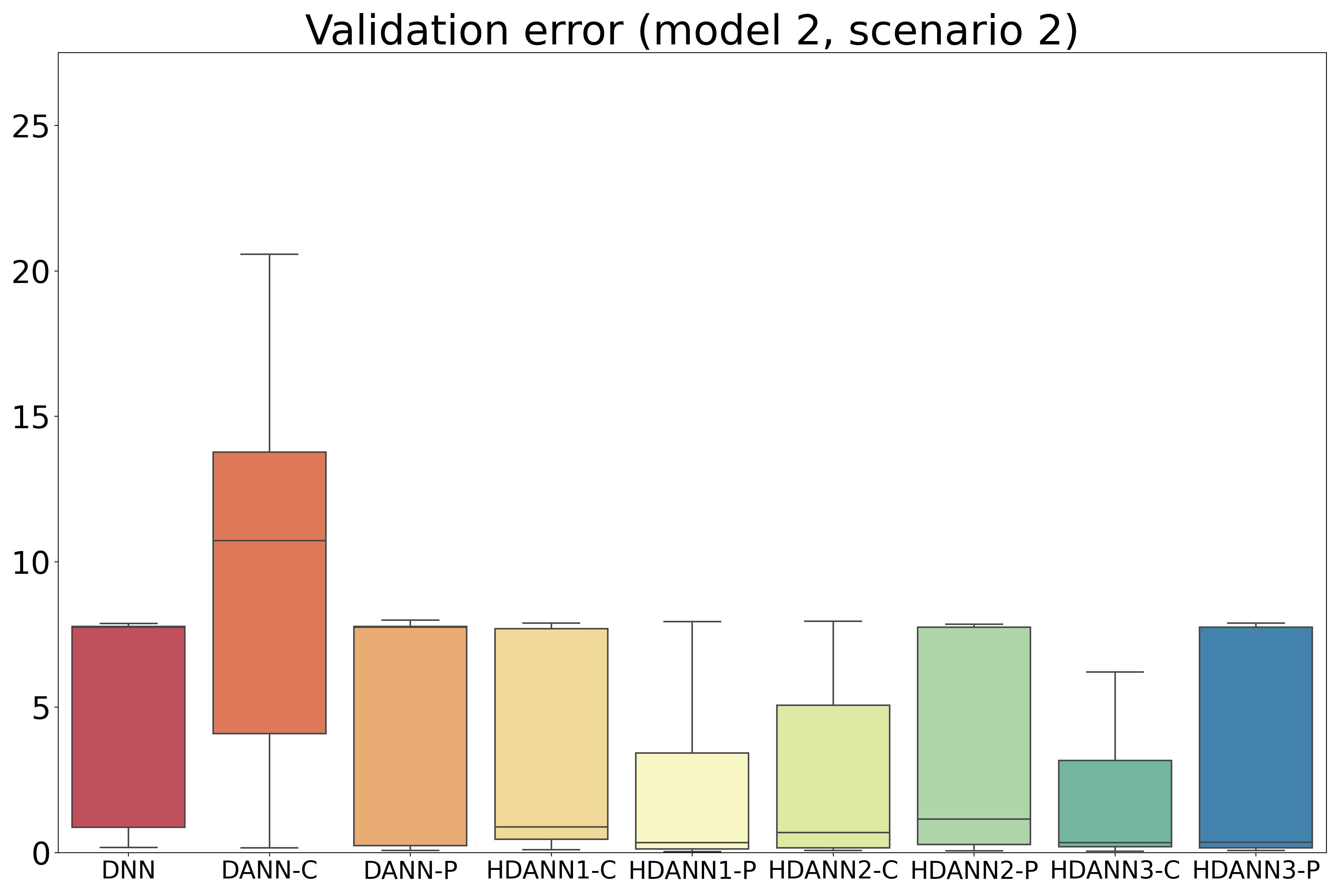}
    \caption{Boxplots of $\sum_{i\in I^{(1)}_{\rm validation}}(Y_i-\hat{Y}_i)^2/500$. The median value and the Q3 value are almost overlapped for the DNN and the DANN with polynomial basis (DANN-P) in some scenarios.}
    \label{fig:boxplot}
\end{figure}

After training the networks, we evaluated validation errors in the original response scale. Figure \ref{fig:boxplot} displays the boxplots of $\sum_{i\in I^{(1)}_{\rm validation}}(Y_i-\hat{Y}_i)^2/500$ across all hyperparameter combinations in the first Monte-Carlo sample, where $I^{(1)}_{\rm validation}$ is the index set of the validation set in the first Monte-Carlo sample. We separated the boxplots for different basis types to see their effects. This figure shows that the DANN with cosine basis (DANN-C) generally yields higher validation errors compared to the DNN, while the other proposed networks tend to have lower validation errors than the DNN. This figure also reveals that the polynomial basis generally yields better performance than the cosine basis for the DANN and HDANN1, while the opposite trend holds for the HDANN2.

\begin{figure}[!ht]
    \centering
    \includegraphics[width=0.475\textwidth]{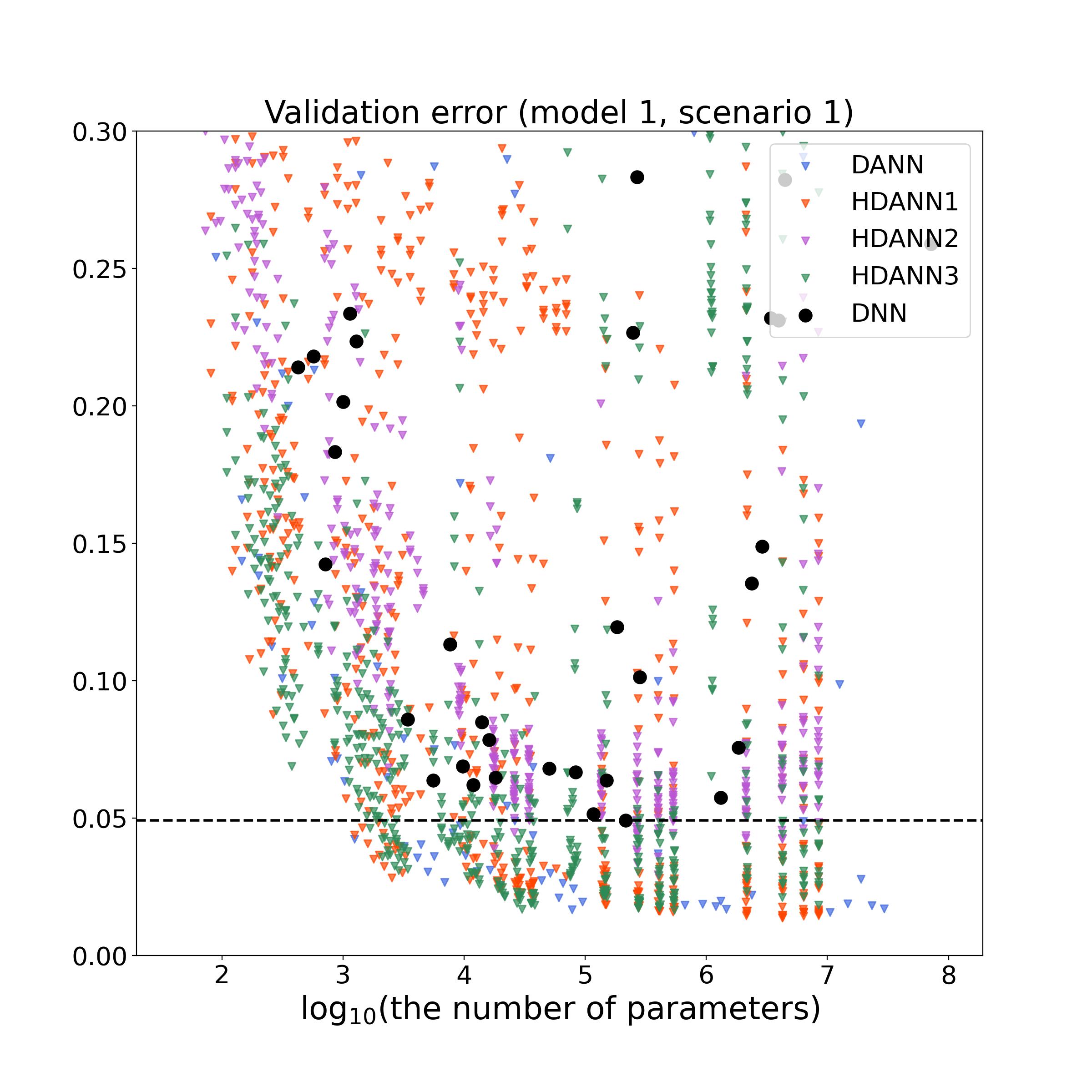}
    \includegraphics[width=0.475\textwidth]{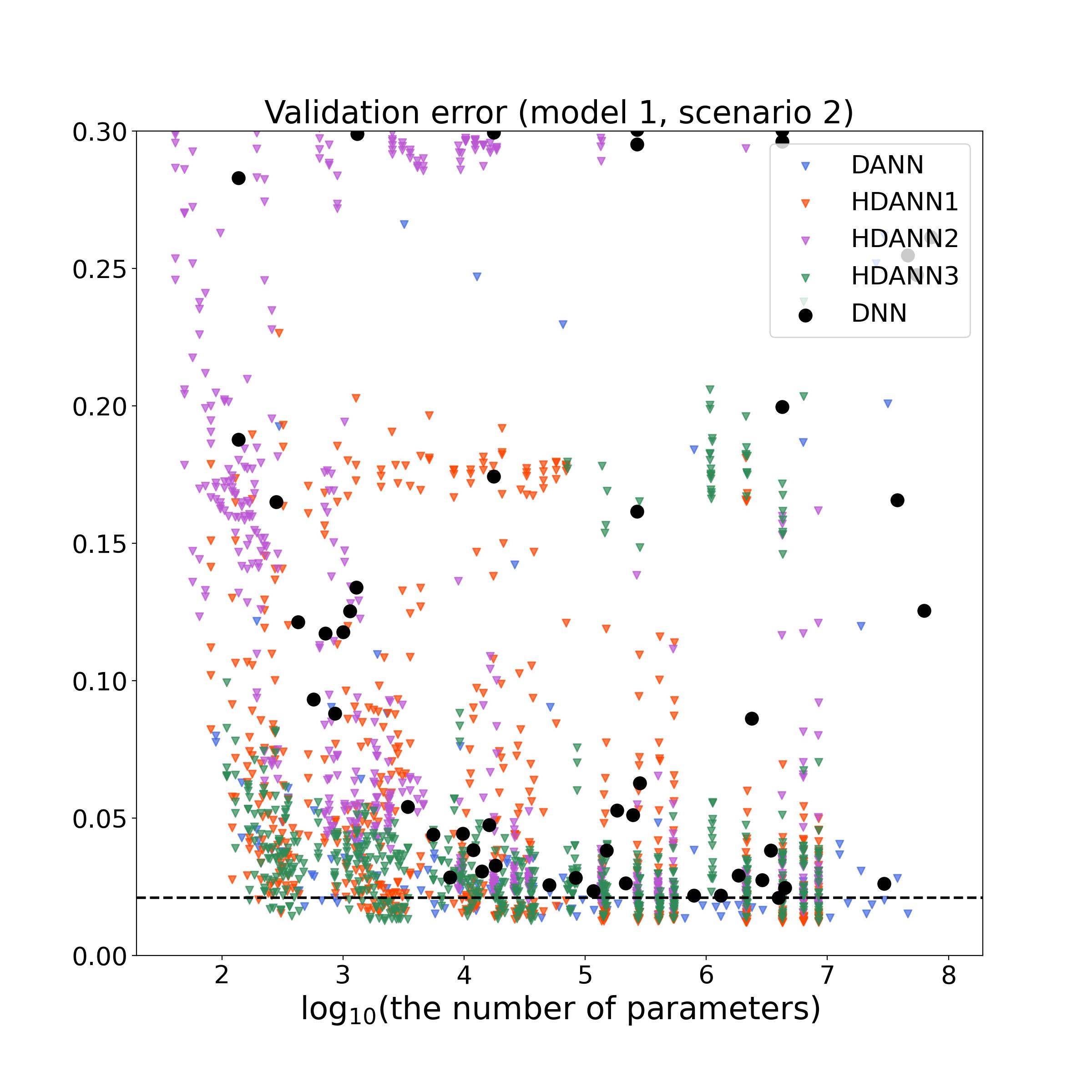}
    \includegraphics[width=0.475\textwidth]{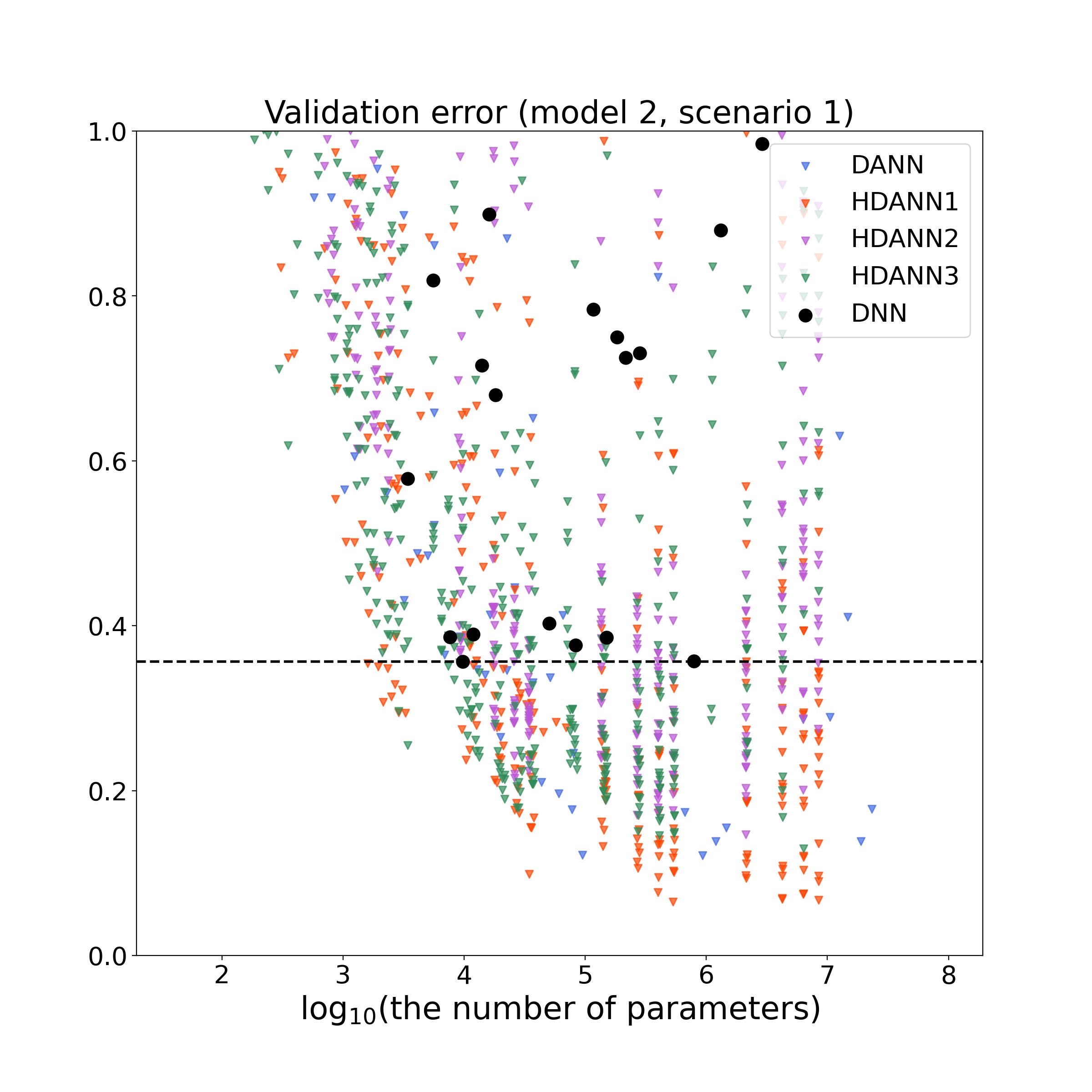}
    \includegraphics[width=0.475\textwidth]{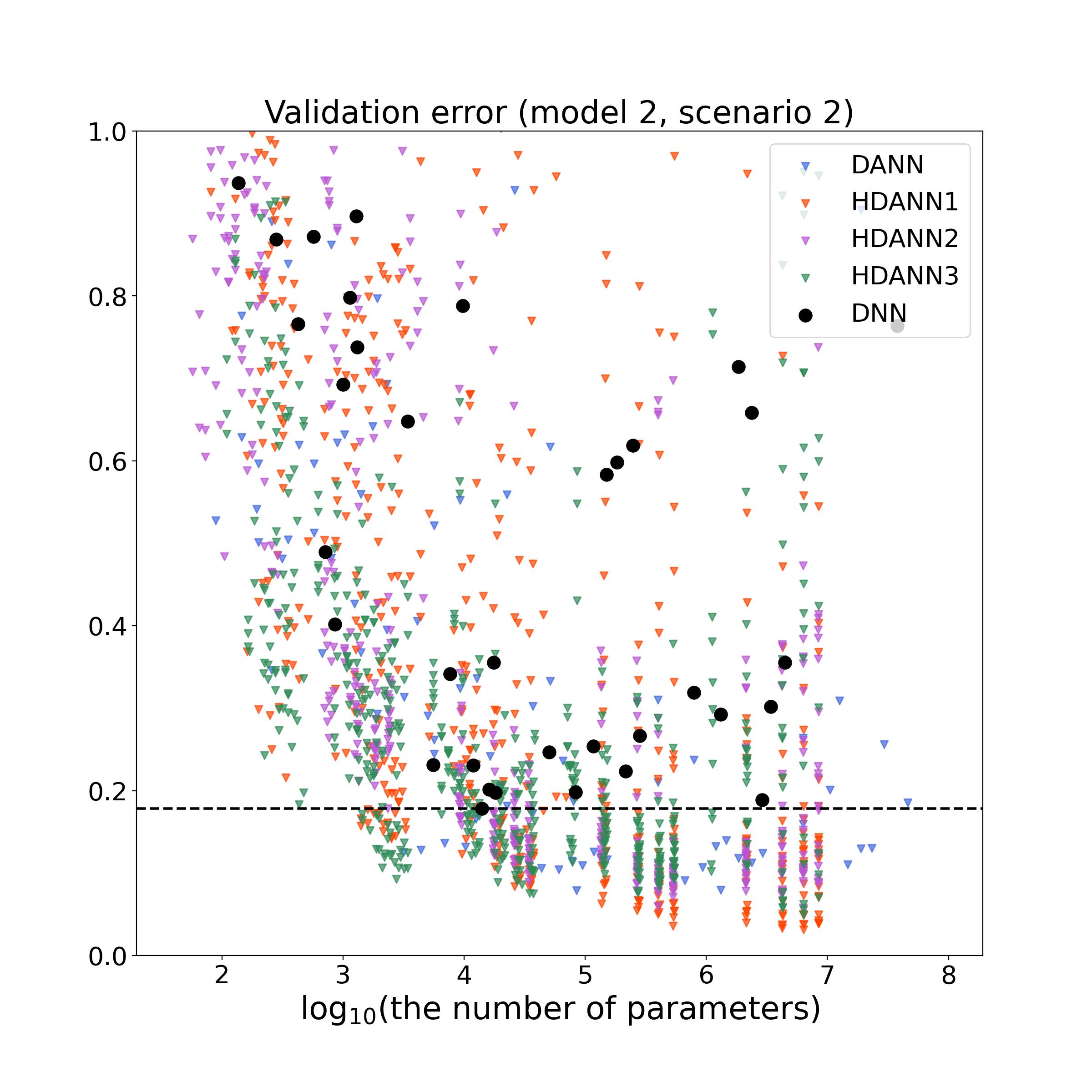}
    \caption{Scatter plots of $\log_{10}(\text{the number of parameters})$ versus $\sum_{i\in I^{(1)}_{\rm validation}}(Y_i-\hat{Y}_i)^2/500$. The black dashed lines represent the smallest validation errors achieved by the DNN.}
    \label{fig:scatter}
\end{figure}

Figure \ref{fig:scatter} illustrates the scatter plots of 
\begin{align*}
\left(\log_{10}(\text{the number of parameters}),\sum_{i\in I^{(1)}_{\rm validation}}(Y_i-\hat{Y}_i)^2/500\right)
\end{align*}
across hyperparameter combinations giving relatively low validation errors in the first Monte-Carlo sample. This figure shows that there always exist proposed networks that achieve lower validation errors than the DNN with much fewer parameters. This demonstrates that adopting an additive layer can significantly reduce network sizes in deep learning.

We also compared the average test errors defined by $5^{-1}\sum_{r=1}^5\sum_{i\in I_{\rm test}^{(r)}}(Y^{(r)}_i-\hat{Y}^{(r)}_i)^2/500$, where $I_{\rm test}^{(r)}$ is the index set of the test set in the $r$th Monte-Carlo sample, $Y^{(r)}_i$ are the response values in that test set and $\hat{Y}^{(r)}_i$ are their predicted values. In the comparison, we only selected either one or two hyperparameter combinations for each network and Monte-Carlo sample. For the DNN, we chose the one giving the lowest validation error in each Monte-Carlo sample. For each proposed network, we chose not only the one giving the lowest validation error but also the one having the smallest number of parameters among hyperparameter combinations giving lower validation errors than the best-tuned DNN in each Monte-Carlo sample. The selected hyperparameter combinations are presented in Appendix \ref{appA}. Table \ref{tab:simulation_result} shows that the best-tuned proposed networks significantly outperform the best-tuned DNN, with HDANN1 performing the best. Table \ref{tab:simulation_result} also shows that the error margins between them increase as the sample size increases. It also shows that the proposed networks equipped with small numbers of parameters have similar performance to the best-tuned DNN. In addition, Appendix \ref{appA} reveals that our networks have similar training times to the DNN. These results demonstrate that our networks are good alternatives to the DNN.

\begin{table}[!ht]
\caption{Comparison of average test errors and average numbers of parameters (in parentheses). The networks hyphenated `best' indicate the best-tuned networks and the networks hyphenated `small' indicate those with the smallest numbers of parameters among hyperparameter combinations giving lower validation errors than the best-tuned DNN.}
\centering
\label{tab:simulation_result}
\footnotesize
\begin{tabular}{@{}rl|rl|rl|rl|rl@{}}
\hline
& & \multicolumn{4}{c|}{Model 1} & \multicolumn{4}{c}{Model 2} \\ \hline
\multicolumn{2}{c|}{Network} & \multicolumn{2}{c|}{Scenario 1} & \multicolumn{2}{c|}{Scenario 2} & \multicolumn{2}{c|}{Scenario 1} & \multicolumn{2}{c}{Scenario 2} \\ \hline
DNN&\hspace{-1.2em}-best      & 0.03898&\hspace{-1.2em} (169159.4)            & 0.02293&\hspace{-1.2em} (901889.0)            & 0.31219&\hspace{-1.2em} (32820.2)             & 0.05285&\hspace{-1.2em} (168737.0) \\
DANN&\hspace{-1.2em}-best     & 0.01588&\hspace{-1.2em} (8837837.8)           & 0.01466&\hspace{-1.2em} (8847361.0)           & 0.11919&\hspace{-1.2em} (816845.8)            & 0.02729&\hspace{-1.2em} (4388353.0) \\
DANN&\hspace{-1.2em}-small    & 0.03298&\hspace{-1.2em} (1876.2)              & 0.02006&\hspace{-1.2em} (11602.6)             & 0.26156&\hspace{-1.2em} (13106.6)             & 0.04165&\hspace{-1.2em} (2107450.6) \\
HDANN1&\hspace{-1.2em}-best   & \textbf{0.01372}&\hspace{-1.2em} (4663297.0)  & \textbf{0.01202}&\hspace{-1.2em} (3813786.6)  & \textbf{0.05652}&\hspace{-1.2em} (5170586.6)  & \textbf{0.01601}&\hspace{-1.2em} (5498061.8) \\
HDANN1&\hspace{-1.2em}-small  & 0.03574&\hspace{-1.2em} (1466.6)              & 0.02064&\hspace{-1.2em} (4604.2)              & 0.27148&\hspace{-1.2em} (1639.4)              & 0.04891&\hspace{-1.2em} (8090.6) \\
HDANN2&\hspace{-1.2em}-best   & 0.02820&\hspace{-1.2em} (1160909.8)           & 0.01664&\hspace{-1.2em} (424858.6)            & 0.12950&\hspace{-1.2em} (2610586.6)           & 0.02560&\hspace{-1.2em} (2369434.6)\\
HDANN2&\hspace{-1.2em}-small  & 0.03460&\hspace{-1.2em} (211809.0)            & 0.02192&\hspace{-1.2em} (51137.0)             & 0.27421&\hspace{-1.2em} (17505.0)             & 0.06055&\hspace{-1.2em} (137897.0)\\
HDANN3&\hspace{-1.2em}-best   & 0.01688&\hspace{-1.2em} (1576193.0)           & 0.01340&\hspace{-1.2em} (6768385.0)           & 0.10530&\hspace{-1.2em} (3053978.6)           & 0.02075&\hspace{-1.2em} (3090330.6)\\ 
HDANN3&\hspace{-1.2em}-small  & 0.03544&\hspace{-1.2em} (2273.0)              & 0.02147&\hspace{-1.2em} (1223.4)              & 0.29031&\hspace{-1.2em} (1889.0)              & 0.04992&\hspace{-1.2em} (6636.2)\\ \hline
\end{tabular}
\end{table}

\section{Real Data Analysis}\label{real data}

We applied our methods to the California Housing data obtained from the {\it sklearn} Python library. This dataset consists of 9 variables across $n=20640$ block groups. A block group is a geographical unit containing multiple houses. We took the logarithm of the median house value in each block group as $Y$ and the following other variables as $X_1,\ldots,X_8$: {\it Longitude of block group, Latitude of block group, Median house age, Average number of rooms, Average number of bedrooms, Population, Average number of households and Median household income.} With these notations, our dataset is written as $\{(X_{i1},\ldots,X_{i8},Y_i):1\leq i\leq n\}$.

To compare the networks, we randomly divided $\{1,\ldots,n\}$ into 5 partitions, say $S^{(1)}, \ldots , S^{(5)}$. We then computed the 5-fold average test errors defined by $5^{-1}\sum_{r=1}^5\sum_{i\in S^{(r)}}(Y_i-\hat{Y}_i)^2/|S^{(r)}|$, where $\hat{Y}_i$ for $i\in S^{(r)}$ is the prediction of $Y_i$ obtained without $\{Y_i:i\in S^{(r)}\}$. To obtain $\hat{Y}_i$ for $i\in S^{(r)}$, we first divided $\{1,\ldots,n\}\setminus S^{(r)}$ into a training set index ($S^{(r)}_{\rm train}$) and a validation set index ($S^{(r)}_{\rm validation}$) with the ratio $3:1$ and scaled all $X_{ij}$ by $(X_{ij}-X^{(r)}_{j,{\rm min}})/(X^{(r)}_{j,{\rm max}}-X^{(r)}_{j,{\rm min}})$, where $X^{(r)}_{j,{\rm min}}$ and $X^{(r)}_{j,{\rm max}}$ are the the minimum and maximum values of $\{X_{ij}:i\in S^{(r)}_{\rm train}\}$, respectively. We then trained the networks using the standardized values of $Y$ with mean and standard deviation being computed from $\{Y_i:i\in S^{(r)}_{\rm train}\}$. Here, we took the same hyperparameter combinations and optimization as in the simulation study. 

\begin{figure}[!ht]
    \centering
    \includegraphics[width=0.6\textwidth]{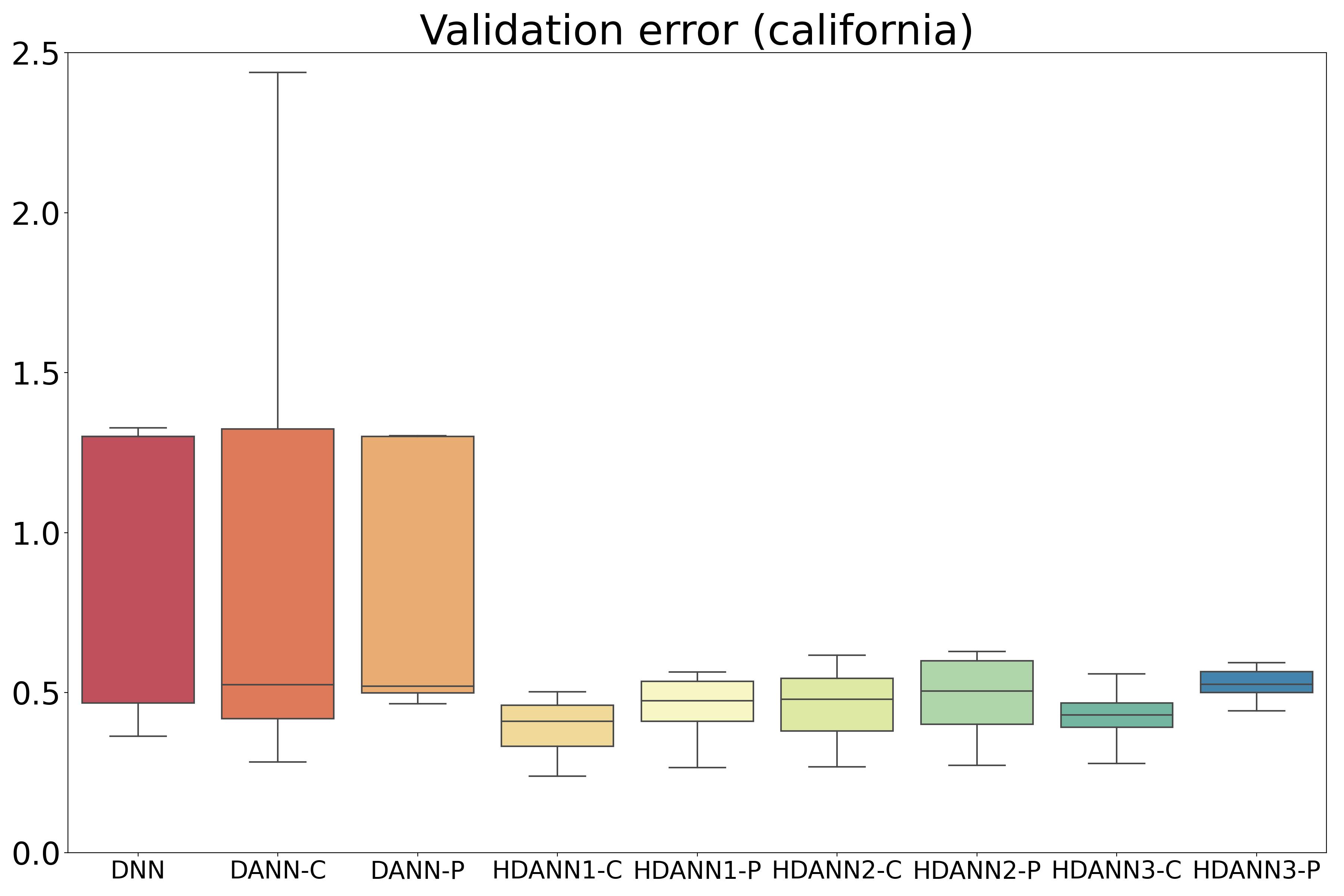}
    \caption{Boxplots of $\sum_{i\in S^{(1)}_{\rm validation}}(Y_i-\hat{Y}_i)^2/|S^{(1)}_{\rm validation}|$. The DNN has almost the same median and Q3 values.}
    \label{fig:Real_boxplot}
\end{figure}

Figure \ref{fig:Real_boxplot} provides the boxplots of validation errors on $S^{(1)}_{\rm validation}$ in the original response scale. It shows that the DANN is comparable with the DNN, while the hybrid networks tend to have significantly lower validation errors than the DNN. In particular, the cosine basis works better than the polynomial basis for the hybrid networks.

\begin{figure}[!ht]
    \centering
    \includegraphics[width=0.6\textwidth]{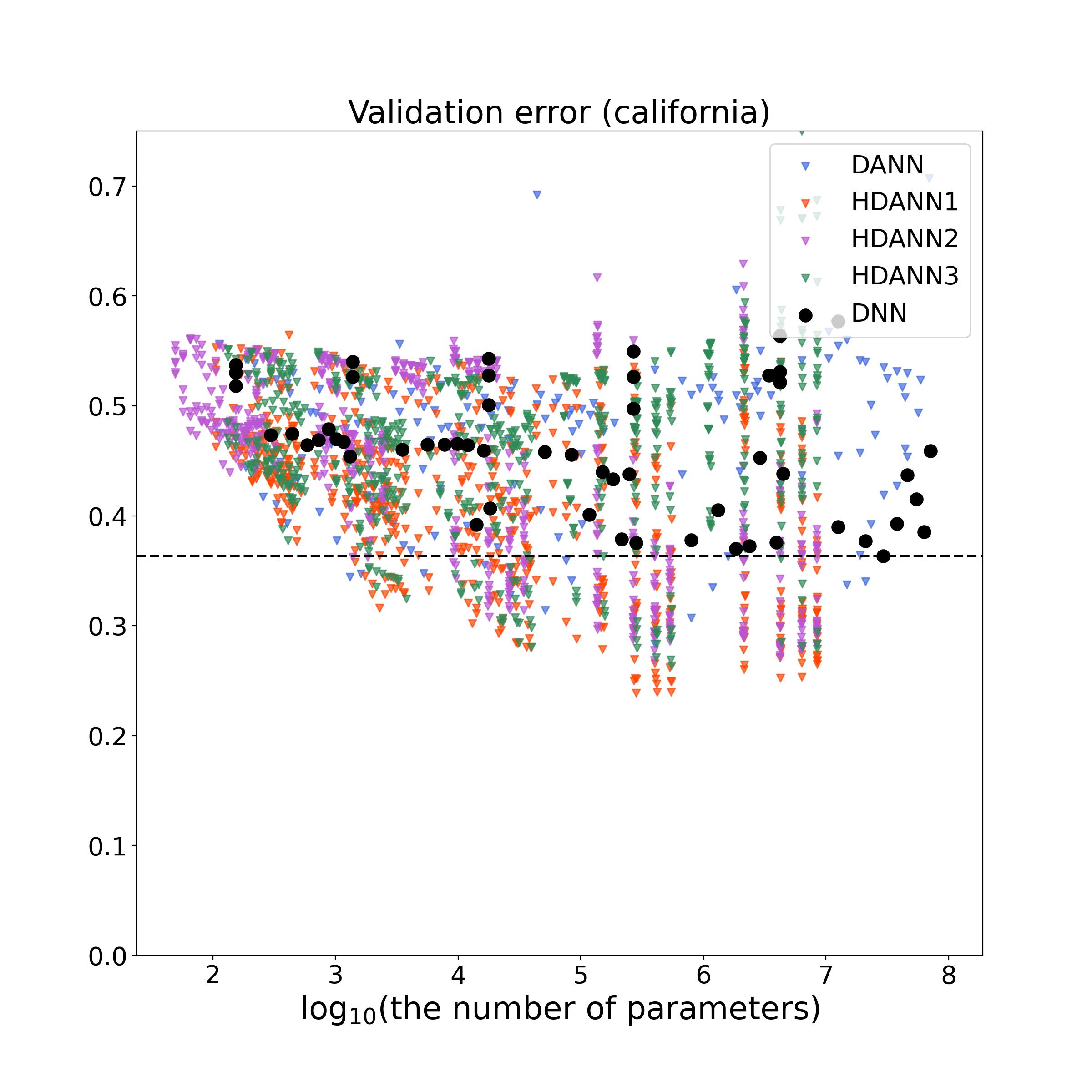}
    \caption{Scatter plot of $\log_{10}(\text{the number of parameters})$ versus $\sum_{i\in S^{(1)}_{\rm validation}}(Y_i-\hat{Y}_i)^2/|S^{(1)}_{\rm validation}|$. The black dashed line represents the smallest validation error achieved by the DNN.}
    \label{fig:Real_scatter}
\end{figure}

To compare the numbers of parameters, the scatter plot of 
\begin{align*}
\left(\log_{10}(\text{the number of parameters}),\sum_{i\in S^{(1)}_{\rm validation}}(Y_i-\hat{Y}_i)^2/|S^{(1)}_{\rm validation}|\right)
\end{align*}
is drawn across hyperparameter combinations with relatively low validation errors (Figure \ref{fig:Real_scatter}). This figure illustrates that there are many proposed networks that have lower validation errors than the DNN. There also exist proposed networks that have approximately $10^4$-times fewer parameters but achieve lower validation errors than the DNN.

In the comparison of $5^{-1}\sum_{r=1}^5\sum_{i\in S^{(r)}}(Y_i-\hat{Y}_i)^2/|S^{(r)}|$, we only considered either one or two hyperparameter combinations for each network based on the errors on $S^{(r)}_{\rm validation}$ similarly as in the simulation study. The selected hyperparameter combinations and the corresponding training times are provided in Appendix \ref{appB}. Table \ref{tab:data analysis_result} shows that all the best-tuned proposed networks significantly outperform the best-tuned DNN, with HDANN1 performing the best. In addition, the proposed networks equipped with small numbers of parameters have lower test errors than the best-tuned DNN. For example, HDANN1-small in Table \ref{tab:data analysis_result} has nearly $10^4$-times fewer parameters than the best-tuned DNN in average. These results again confirm the superiority of our networks.

\begin{table}[!ht]
\caption{Comparison of average test errors and average numbers of parameters (in parentheses) on the California Housing data.}
\centering
\label{tab:data analysis_result}
\begin{tabular}{rl|rl}
\hline
\multicolumn{2}{c|}{Network} & \multicolumn{2}{c}{Test Error (\# of Parameters)} \\ \hline
DNN&\hspace{-1em}-best     & \hspace{2.5em}0.38249& \hspace{-1.2em} (8194458.6) \\
DANN&\hspace{-1em}-best    & 0.30345&\hspace{-1.2em} (334823.4) \\
DANN&\hspace{-1em}-small   & 0.36606&\hspace{-1.2em} (905.0) \\
HDANN1&\hspace{-1em}-best  & \textbf{0.24706}&\hspace{-1.2em} (332289.0) \\
HDANN1&\hspace{-1em}-small & 0.37593&\hspace{-1.2em} (794.6) \\
HDANN2&\hspace{-1em}-best  & 0.28298&\hspace{-1.2em} (5549159.4) \\
HDANN2&\hspace{-1em}-small & 0.36084&\hspace{-1.2em} (5601.0) \\
HDANN3&\hspace{-1em}-best  & 0.27396&\hspace{-1.2em} (493773.8) \\
HDANN3&\hspace{-1em}-small & 0.37514&\hspace{-1.2em} (1217.8) \\ \hline
\end{tabular}
\end{table}

\section{Conclusion}\label{conclusion}

In this works, we proposed a new nonlinear deep neural network along with hybrid variants. These networks are easy to implement due to the simple construction of activation and basis functions. Additionally, these networks achieve the universal approximation properties. We found that adding nonlinear layers enhances accuracy and can even reduce the number of parameters. The idea of this work can be applied to other topics, such as convolutional neural networks and classification problems, by making appropriate changes. We belive that our networks are promissing alternatives to the traditional neural networks.

\section{Appendix}

\subsection{Proofs}

\subsubsection{Proof of Lemma \ref{lem1}}

The desired result holds for $B_r(x)=x^r$ since the space of all polynomials on $[0,1]$ is a dense subset of the space of all continuous functions on $[0,1]$ by the Weierstrass approximation theorem.

The desired result also holds for $B_r(x)=\cos(r\pi x)$. To see this, define a function $\Phi:[-\pi,\pi]\rightarrow\mathbb{R}$ by $\Phi(z)=\phi(z/\pi)\I(z\in[0,\pi])+\phi(-z/\pi)\I(z\in[-\pi,0])$. Then, $\Phi$ is an even continuous function, and hence Fej\'{e}r's theorem (e.g., Theorem 4.32 of \cite{Pereyra and Ward (2012)}) implies that a function $\sigma_N\Phi$ defined by 
\begin{align*}
\sigma_N\Phi(z)=\frac{1}{2\pi(N+1)}\sum_{k=0}^N\sum_{s=-k}^k\int_{-\pi}^{\pi}\Phi(t)(\cos(st)-\sqrt{-1}\sin(st))dt(\cos(sz)+\sqrt{-1}\sin(sz))
\end{align*}
converges uniformly to $\Phi$. Since $\sin$ is an odd function, we have $\int_{-\pi}^{\pi}\Phi(t)\sin(st)dt=0$. This with the fact that $\Phi$ is real-valued implies that
\begin{align*}
\sigma_N\Phi(z)=\frac{1}{2\pi(N+1)}\sum_{k=0}^N\sum_{s=-k}^k\int_{-\pi}^{\pi}\Phi(t)\cos(st)dt\cos(sz)
\end{align*}
converges uniformly to $\Phi$. Note that $\sigma_N\Phi(z)$ is a linear combination of $\{\cos(sx):0\leq s\leq N\}$ since $\cos$ is an even function. Hence, for given $\epsilon>0$, there exists $N\geq0$ and $c_s\in\mathbb{R}$ such that $\sup_{z\in[0,\pi]}|\Phi(z)-\sum_{s=0}^Nc_s\cos(sz)|<\epsilon$. This implies that $\sup_{x\in[0,1]}|\phi(x)-\sum_{s=0}^Nc_s\cos(s\pi x)|<\epsilon$.

Lastly, it is well known that the desired result holds for the Haar basis on $[0,1]$; see \cite{McLaughlin (1969)}, for example. \qed

\subsubsection{Proof of Theorem \ref{thm1}}

The universal approximation theorem (\cite{Cybenko (1989)}) implies that there exist constants $p\geq1$ and $w_k,w_{jk},a_k\in\mathbb{R}$ such that 
\begin{align*}
\sup_{\mx\in[0,1]^d}\left|f(\mx)-\left(\sum_{k=1}^p w_kg\left(\sum_{j=1}^d w_{jk}x_j + a_k\right)\right)\right|<\frac{\epsilon}{2}.
\end{align*}
Note that the functions $h_k:x\mapsto w_kx$ and $h_{jk}:x_j\mapsto w_{jk}x_j+a_k/d$ are uniformly continuous on $[0,1]$. Hence, there exists $\delta_k>0$ such that 
\begin{align*}
\sup_{x,y\in[0,1]:|x-y|<\delta_k}|h_k(x)-h_k(y)|<\frac{\epsilon}{4p}. 
\end{align*}
By the definitions of $B_{jkr}$ and $B_{k\ell}$, there exist constants $q_k, q_{jk}\geq1$ and $c_{k\ell}, b_k, c_{jkr}, b_{jk}\in\mathbb{R}$ such that 
\begin{align*}
\sup_{x\in[0,1]}\left|h_k(x)-\left(\sum_{\ell=1}^{q_k}c_{k\ell}B_{k\ell}(x)+b_k\right)\right|&<\frac{\epsilon}{4p},\\
\sup_{x_j\in[0,1]}\left|h_{jk}(x_j)-\left(\sum_{r=1}^{q_{jk}}c_{jkr}B_{jkr}(x_j)+b_{jk}\right)\right|&<\frac{\delta_k}{dL},
\end{align*}
where $L>0$ is the Lipschitz constant of $g$. Define 
\begin{align*}
\tilde{h}_k(x)&=\sum_{\ell=1}^{q_k}c_{k\ell}B_{k\ell}(x)+b_k,\\
\tilde{h}_{jk}(x_j)&=\sum_{r=1}^{q_{jk}}c_{jkr}B_{jkr}(x_j)+b_{jk}.
\end{align*}
Note that
\begin{align*}
\left|g\left(\sum_{j=1}^dh_{jk}(x_j)\right)-g\left(\sum_{j=1}^d\tilde{h}_{jk}(x_j)\right)\right|&\leq L\left|\sum_{j=1}^dh_{jk}(x_j)-\sum_{j=1}^d\tilde{h}_{jk}(x_j)\right|\\
&\leq L\sum_{j=1}^d|h_{jk}(x_j)-\tilde{h}_{jk}(x_j)|\\
&\leq\delta_k.
\end{align*}
Then,
\begin{align*}
&\sup_{\mx\in[0,1]^d}\left|\sum_{k=1}^ph_k\left(g\left(\sum_{j=1}^dh_{jk}(x_j)\right)\right)-\sum_{k=1}^p\tilde{h}_k\left(g\left(\sum_{j=1}^d\tilde{h}_{jk}(x_j)\right)\right)\right|\\
&\leq\sup_{\mx\in[0,1]^d}\sum_{k=1}^p\left|h_k\left(g\left(\sum_{j=1}^dh_{jk}(x_j)\right)\right)-\tilde{h}_k\left(g\left(\sum_{j=1}^d\tilde{h}_{jk}(x_j)\right)\right)\right|\\
&\leq\sup_{\mx\in[0,1]^d}\sum_{k=1}^p\left|h_k\left(g\left(\sum_{j=1}^dh_{jk}(x_j)\right)\right)-h_k\left(g\left(\sum_{j=1}^d\tilde{h}_{jk}(x_j)\right)\right)\right|\\
&\quad+\sup_{\mx\in[0,1]^d}\sum_{k=1}^p\left|h_k\left(g\left(\sum_{j=1}^d\tilde{h}_{jk}(x_j)\right)\right)-\tilde{h}_k\left(g\left(\sum_{j=1}^d\tilde{h}_{jk}(x_j)\right)\right)\right|\\
&\leq\sum_{k=1}^p\sup_{\mx\in[0,1]^d}\left|h_k\left(g\left(\sum_{j=1}^dh_{jk}(x_j)\right)\right)-h_k\left(g\left(\sum_{j=1}^d\tilde{h}_{jk}(x_j)\right)\right)\right|\\
&\quad+\sum_{k=1}^p\sup_{\mx\in[0,1]^d}\left|h_k\left(g\left(\sum_{j=1}^d\tilde{h}_{jk}(x_j)\right)\right)-\tilde{h}_k\left(g\left(\sum_{j=1}^d\tilde{h}_{jk}(x_j)\right)\right)\right|\\
&\leq\sum_{k=1}^p\sup_{x,y\in[0,1]:|x-y|<\delta_k}|h_k(x)-h_k(y)|+\sum_{k=1}^p\sup_{x\in[0,1]}|h_k(x)-\tilde{h}_k(x)|\\
&\leq\frac{\epsilon}{4}+\frac{\epsilon}{4}.
\end{align*}
Therefore,
\begin{align*}
&\sup_{\mx\in[0,1]^d}\left|f(\mx)-\sum_{k=1}^p\tilde{h}_k\left(g\left(\sum_{j=1}^d\tilde{h}_{jk}(x_j)\right)\right)\right|\\
&\leq\sup_{\mx\in[0,1]^d}\left|f(\mx)-\sum_{k=1}^ph_k\left(g\left(\sum_{j=1}^dh_{jk}(x_j)\right)\right)\right|\\
&\quad+\sup_{\mx\in[0,1]^d}\left|\sum_{k=1}^ph_k\left(g\left(\sum_{j=1}^dh_{jk}(x_j)\right)\right)-\sum_{k=1}^p\tilde{h}_k\left(g\left(\sum_{j=1}^d\tilde{h}_{jk}(x_j)\right)\right)\right|\\
&<\epsilon.
\end{align*}

\subsubsection{Proof of Theorem \ref{thm2}}

Write $\sigma_1=\cdots=\sigma_L=\sigma$. By Theorem 3.2 of \cite{Kidger and Lyons (2020)}, there exist $L\geq1$, $W_k^{(1)}\in\mathbb{R}^d$, $W_k^{(2)},\ldots,W_k^{(L)}\in\mathbb{R}^{d+3}$ and $w_k, a_k^{(1)}, b_k^{(l)}, b\in\mathbb{R}$ such that
\begin{align*}
\sup_{\mx\in[0,1]^d}\left|f(\mx)-\left(\sum_{k=1}^{d+3}w_k \tilde{x}_{k}^{(L)}+b\right)\right|<\frac{\epsilon}{2},
\end{align*}
where
\begin{align*}
\tilde{x}_k^{(l)} & = \sigma\left({W_k^{(l)}}\tran\tilde{\mx}^{(l-1)}+b_k^{(l)}\right)
\end{align*}
for $1\leq k\leq d+3$ and $2\leq l\leq L$, and
\begin{align*}
\tilde{x}_k^{(1)} & = \sigma\left({W_k^{(1)}}\tran\mx+a_k^{(1)}\right)
\end{align*}
for $1\leq k\leq d+3$. Define $M=\max\{|w_1|,\ldots,|w_{d+3}|\}$ and $M_{kl}=\|W_k^{(l)}\|_\infty$. Let $C>0$ denote the Lipschitz constant of $\sigma$. Define $h_{jk}:[0,1]\rightarrow\mathbb{R}$ by $h_{jk}(x_j)=W_{kj}^{(1)}x_j+a_k^{(1)}/d$, where $W_{kj}^{(1)}$ is the $j$th entry of $W_k^{(1)}$. Then, $h_{jk}$ is a continuous function on $[0,1]$. By Lemma \ref{lem1}, there exist constants $q_{jk}\geq1$ and $c_{jkr}, b_{jk}\in\mathbb{R}$ such that
\begin{align*}
\sup_{x_j\in[0,1]}\left|h_{jk}(x_j)-\left(\sum_{r=1}^{q_{jk}}c_{jkr}B_{jkr}(x_j)+b_{jk}\right)\right|<\frac{\epsilon}{2 M C^L \left(\max_{1\leq k\leq d+3, 2\leq l\leq L}M_{kl}\right) (d+3)^{L} d}.
\end{align*}
Define $\bar{h}_{jk}(x_j)=\sum_{r=1}^{q_{jk}}c_{jkr}B_{jkr}(x_j)+b_{jk}$. Then,
\begin{align*}
&\sup_{\mx\in[0,1]^d}\left|\sigma\left({W_k^{(1)}}\tran\mx+a_k^{(1)}\right)-\sigma\left(\sum_{j=1}^d\bar{h}_{jk}(x_j)\right)\right|\\
&\leq C\sup_{\mx\in[0,1]^d}\left|\sum_{j=1}^dh_{jk}(x_j)-\sum_{j=1}^d\bar{h}_{jk}(x_j)\right|\\
&\leq C\sup_{\mx\in[0,1]^d}\sum_{j=1}^d|h_{jk}(x_j)-\bar{h}_{jk}(x_j)|\\
&\leq \frac{\epsilon}{2 M C^{L-1} \left(\max_{1\leq k\leq d+3, 2\leq l\leq L}M_{kl}\right) (d+3)^{L}}
\end{align*}
Define
\begin{align*}
x_k^{(l)} & = \sigma\left({W_k^{(l)}}\tran\mx^{(l-1)}+b_k^{(l)}\right)
\end{align*}
for $1\leq k\leq d+3$ and $2\leq l\leq L$, and
\begin{align*}
x_k^{(1)} & = \sigma\left(\sum_{j=1}^d\bar{h}_{jk}(x_j)\right).
\end{align*}
Since
\begin{align*}
\left|\sigma\left({W_k^{(l)}}\tran\tilde{\mx}^{(l-1)}+b_k^{(l)}\right)-\sigma\left({W_k^{(l)}}\tran\mx^{(l-1)}+b_k^{(l)}\right)\right|&\leq C|{W_k^{(l)}}\tran(\tilde{\mx}^{(l-1)}-\mx^{(l-1)})|\\
&\leq C M_{kl} \sum_{s=1}^{d+3}|\tilde{x}^{(l-1)}_s-x^{(l-1)}_s|
\end{align*}
for $2\leq l\leq L$, we get
\begin{align*}
&\sup_{\mx\in[0,1]^d} \left|\sum_{k=1}^{d+3}w_k \tilde{x}_{k}^{(L)}+b-\left(\sum_{k=1}^{d+3}w_k x_{k}^{(L)}+b\right)\right|\\
&\leq \sup_{\mx\in[0,1]^d} M\sum_{k=1}^{d+3}|\tilde{x}_{k}^{(L)}-x_{k}^{(L)}|\\
&\leq \sup_{\mx\in[0,1]^d} M C \sum_{k=1}^{d+3} M_{kL} \sum_{s=1}^{d+3}|\tilde{x}^{(L-1)}_s-x^{(L-1)}_s|\\
&\leq \cdots\\
&\leq \sup_{\mx\in[0,1]^d} M C^{L-1} \left(\max_{1\leq k\leq d+3, 2\leq l\leq L}M_{kl}\right) (d+3)^{L-1} \sum_{s=1}^{d+3}|\tilde{x}^{(1)}_s-x^{(1)}_s|\\
&\leq \frac{\epsilon}{2}.
\end{align*}
Therefore,
\begin{align*}
\sup_{\mx\in[0,1]^d}\left|f(\mx)-\left(\sum_{k=1}^{d+3}w_kx_{k}^{(L)}+b\right)\right|<\epsilon.
\end{align*}

\subsubsection{Proof of Theorem \ref{thm3}}

Write $\sigma_1=\cdots=\sigma_{L-1}=g_L=\sigma$. By Theorem 3.2 of \cite{Kidger and Lyons (2020)}, there exist $L\geq1$, $W_k^{(1)}\in\mathbb{R}^d$, $W_k^{(2)},\ldots,W_k^{(L)}\in\mathbb{R}^{d+3}$ and $w_k, b_k^{(l)}, a\in\mathbb{R}$ such that
\begin{align*}
\sup_{\mx\in[0,1]^d}\left|f(\mx)-\left(\sum_{k=1}^{d+3}w_k x_{k}^{(L)}+a\right)\right|<\frac{\epsilon}{2},
\end{align*}
where
\begin{align*}
x_k^{(l)} & = \sigma\left({W_k^{(l)}}\tran\mx^{(l-1)}+b_k^{(l)}\right)
\end{align*}
for $1\leq k\leq d+3$ and $2\leq l\leq L$, and
\begin{align*}
x_k^{(1)} & = \sigma\left({W_k^{(1)}}\tran\mx+b_k^{(1)}\right)
\end{align*}
for $1\leq k\leq d+3$. Define $h_k:[0,1]\rightarrow\mathbb{R}$ by $h_k(x_k)=w_k x_k+a/(d+3)$. Then, $h_k$ is a continuous function on $[0,1]$. By Lemma \ref{lem1}, there exist constants $q_k\geq1$ and $c_{kr}, b_k\in\mathbb{R}$ such that
\begin{align*}
\sup_{x_k\in[0,1]}\left|h_k(x_k)-\left(\sum_{r=1}^{q_{k}}c_{kr}B_{kr}(x_k)+b_{k}\right)\right|<\frac{\epsilon}{2(d+3)}.
\end{align*}
Then,
\begin{align*}
\sup_{\mx\in[0,1]^d}\left|\sum_{k=1}^{d+3}h_k(x_k)-\sum_{k=1}^{d+3}\left(\sum_{r=1}^{q_{k}}c_{kr}B_{kr}(x_k)+b_{k}\right)\right|<\frac{\epsilon}{2}.
\end{align*}
Therefore,
\begin{align*}
\sup_{\mx\in[0,1]^d}\left|f(\mx)-\sum_{k=1}^{d+3}\left(\sum_{r=1}^{q_{k}}c_{kr}B_{kr}(x_{k}^{(L)})+b_{k}\right)\right|<\epsilon.
\end{align*}



\clearpage

\subsection{Selected hyperparameters for Table \ref{tab:simulation_result}}\label{appA}

\begin{table}[!h]
\caption{Model 1 in scenario 1}
\centering
\label{tab:Exponential_small_monte_top_5}
\tiny
\begin{tabular}{c|rl|ccccc|cccc}
\hline
Monte-Carlo sample & \multicolumn{2}{c|}{Network} & $L$ & $p$ & $q$ & $\sigma$ & $B_r$ & Validation Error & Test Error & Training Time (sec) & \# of Parameters \\ \hline
            & DNN&\hspace{-1.54em}-best             & 14 & 128 & & tanh & & 0.04912 & 0.03219 & 10.13 & 215681 \\ 
            & DANN&\hspace{-1.54em}-best            & 3 & 1024 & 5 & & poly & 0.01568 & 0.01699 & 10.40 & 10524673 \\ 
            & DANN&\hspace{-1.54em}-small     & 1 & 16 & 11 & & poly & 0.04245 & 0.04857 & 10.29 & 1249 \\ 
            & HDANN1&\hspace{-1.54em}-best          & 5 & 1024 & 7 & ReLU & poly & 0.01375 & 0.01594 & 3.84 & 4243457 \\ 
{1st}  & HDANN1&\hspace{-1.54em}-small   & 3 & 16 & 7 & ReLU & poly & 0.04384 & 0.04862 & 9.26 & 1249 \\ 
            & HDANN2&\hspace{-1.54em}-best          & 7 & 256 & 9 & tanh & poly & 0.02946 & 0.01961 & 13.14 & 398849 \\ 
            & HDANN2&\hspace{-1.54em}-small   & 5 & 64 & 9 & tanh & poly & 0.03890 & 0.02693 & 16.46 & 17665 \\ 
            & HDANN3&\hspace{-1.54em}-best          & 9 & 256 & 11 & ReLU & poly & 0.01648 & 0.01751 & 9.65 & 546305 \\ 
            & HDANN3&\hspace{-1.54em}-small   & 5 & 16 & 5 & ReLU & poly & 0.04639 & 0.04658 & 12.87 & 1665 \\ \hline 
            & DNN&\hspace{-1.54em}-best             & 16 & 128 & & tanh & & 0.03601 & 0.04036 & 16.61 & 248705 \\ 
            & DANN&\hspace{-1.54em}-best            & 3 & 256 & 5 & & poly & 0.01563 & 0.01482 & 9.82 & 665089 \\ 
            & DANN&\hspace{-1.54em}-small     & 9 & 4 & 11 & & poly & 0.02473 & 0.02340 & 110.74 & 1753 \\ 
            & HDANN1&\hspace{-1.54em}-best          & 7 & 1024 & 5 & ReLU & poly & 0.01280 & 0.01269 & 5.67 & 6330369 \\ 
{2nd}  & HDANN1&\hspace{-1.54em}-small   & 3 & 16 & 5 & ReLU & poly & 0.03287 & 0.03253 & 13.12 & 1057 \\ 
            & HDANN2&\hspace{-1.54em}-best          & 9 & 256 & 5 & tanh & poly & 0.02912 & 0.02662 & 17.63 & 529409 \\ 
            & HDANN2&\hspace{-1.54em}-small   & 9 & 256 & 5 & tanh & poly & 0.02912 & 0.02662 & 17.63 & 529409 \\ 
            & HDANN3&\hspace{-1.54em}-best          & 9 & 256 & 5 & ReLU & poly & 0.01526 & 0.01437 & 8.36 & 535553 \\ 
            & HDANN3&\hspace{-1.54em}-small   & 5 & 4 & 7 & ReLU & poly & 0.02893 & 0.03243 & 44.98 & 281 \\ \hline 
            & DNN&\hspace{-1.54em}-best             & 12 & 128 & & tanh & & 0.03655 & 0.03968 & 13.85 & 182657 \\ 
            & DANN&\hspace{-1.54em}-best            & 3 & 256 & 11 & & poly & 0.01530 & 0.01723 & 13.32 & 1462273 \\ 
            & DANN&\hspace{-1.54em}-small     & 1 & 16 & 11 & & poly & 0.03546 & 0.04420 & 17.96 & 1249 \\ 
            & HDANN1&\hspace{-1.54em}-best          & 5 & 1024 & 9 & ReLU & poly & 0.01419 & 0.01524 & 5.36 & 4255745 \\ 
{3th}  & HDANN1&\hspace{-1.54em}-small   & 3 & 16 & 7 & ReLU & poly & 0.03300 & 0.03943 & 10.67 & 1249 \\ 
            & HDANN2&\hspace{-1.54em}-best          & 7 & 256 & 5 & tanh & poly & 0.02505 & 0.02811 & 15.29 & 397825 \\ 
            & HDANN2&\hspace{-1.54em}-small   & 9 & 64 & 3 & tanh & cos & 0.03651 & 0.03975 & 15.10 & 33921 \\ 
            & HDANN3&\hspace{-1.54em}-best          & 7 & 256 & 11 & ReLU & poly & 0.01555 & 0.01936 & 10.87 & 414721 \\ 
            & HDANN3&\hspace{-1.54em}-small   & 5 & 16 & 5 & ReLU & poly & 0.03628 & 0.04354 & 17.56 & 1665 \\ \hline 
            & DNN&\hspace{-1.54em}-best             & 12 & 128 & & tanh & & 0.04206 & 0.03322 & 17.76 & 182657 \\ 
            & DANN&\hspace{-1.54em}-best            & 5 & 1024 & 5 & & poly & 0.01798 & 0.01697 & 21.24 & 21012481 \\ 
            & DANN&\hspace{-1.54em}-small     & 1 & 64 & 9 & & poly & 0.03951 & 0.04386 & 17.93 & 4097 \\ 
            & HDANN1&\hspace{-1.54em}-best          & 7 & 1024 & 7 & ReLU & poly & 0.01461 & 0.01387 & 6.80 & 6342657 \\ 
{4th}  & HDANN1&\hspace{-1.54em}-small   & 5 & 16 & 9 & ReLU & poly & 0.03809 & 0.03149 & 13.40 & 1985 \\ 
            & HDANN2&\hspace{-1.54em}-best          & 5 & 256 & 11 & tanh & poly & 0.03115 & 0.02291 & 18.35 & 267777 \\ 
            & HDANN2&\hspace{-1.54em}-small   & 5 & 256 & 5 & tanh & poly & 0.03815 & 0.02401 & 18.90 & 266241 \\ 
            & HDANN3&\hspace{-1.54em}-best          & 7 & 1024 & 7 & ReLU & poly & 0.02388 & 0.02013 & 10.95 & 6348801 \\ 
            & HDANN3&\hspace{-1.54em}-small   & 1 & 64 & 7 & sigmoid & poly & 0.04018 & 0.03666 & 18.00 & 7361 \\ \hline 
            & DNN&\hspace{-1.54em}-best             & 16 & 32 & & tanh & & 0.03117 & 0.04580 & 21.02 & 16097 \\ 
            & DANN&\hspace{-1.54em}-best            & 3 & 1024 & 5 & & poly & 0.01481 & 0.02156 & 13.26 & 10524673 \\ 
            & DANN&\hspace{-1.54em}-small     & 5 & 4 & 11 & & poly & 0.02274 & 0.06149 & 58.08 & 1033 \\ 
            & HDANN1&\hspace{-1.54em}-best          & 3 & 1024 & 7 & ReLU & poly & 0.01323 & 0.01779 & 3.87 & 2144257 \\ 
{5th}  & HDANN1&\hspace{-1.54em}-small   & 5 & 16 & 7 & ReLU & poly & 0.03088 & 0.05785 & 12.18 & 1793 \\ 
            & HDANN2&\hspace{-1.54em}-best          & 5 & 1024 & 5 & ReLU & poly & 0.02621 & 0.03963 & 9.37 & 4210689 \\ 
            & HDANN2&\hspace{-1.54em}-small   & 7 & 64 & 3 & tanh & poly & 0.03031 & 0.03333 & 19.76 & 25601 \\ 
            & HDANN3&\hspace{-1.54em}-best          & 9 & 64 & 5 & ReLU & poly & 0.01324 & 0.03074 & 11.70 & 35585 \\ 
            & HDANN3&\hspace{-1.54em}-small   & 5 & 4 & 11 & ReLU & poly & 0.02543 & 0.08000 & 40.19 & 393 \\ \hline 

\end{tabular}
\end{table}

\begin{table}[!h]
\caption{Model 1 in scenario 2}
\centering
\label{tab:Exponential_big_monte_top_5}
\tiny
\begin{tabular}{c|rl|ccccc|cccc}
\hline
Monte-Carlo sample & \multicolumn{2}{c|}{Network} & $L$ & $p$ & $q$ & $\sigma$ & $B_r$ & Validation Error & Test Error & Training Time (sec) & \# of Parameters \\ \hline
            & DNN&\hspace{-1.54em}-best             & 16 & 512 & & tanh & & 0.02109 & 0.02620 & 12.09 & 3943937 \\ 
            & DANN&\hspace{-1.54em}-best            & 3 & 256 & 5 & & poly & 0.01359 & 0.01867 & 8.99 & 665089 \\ 
            & DANN&\hspace{-1.54em}-small     & 5 & 4 & 5 & & poly & 0.01780 & 0.07074 & 56.97 & 481 \\ 
            & HDANN1&\hspace{-1.54em}-best          & 3 & 1024 & 7 & ReLU & poly & 0.01202 & 0.01208 & 2.94 & 2144257 \\ 
{1st}  & HDANN1&\hspace{-1.54em}-small   & 7 & 4 & 3 & tanh & poly & 0.02038 & 0.07681 & 28.59 & 201 \\ 
            & HDANN2&\hspace{-1.54em}-best          & 7 & 256 & 11 & tanh & poly & 0.01467 & 0.01718 & 12.48 & 399361 \\ 
            & HDANN2&\hspace{-1.54em}-small   & 9 & 64 & 3 & tanh & cos & 0.02015 & 0.02593 & 11.05 & 33921 \\ 
            & HDANN3&\hspace{-1.54em}-best          & 9 & 1024 & 5 & ReLU & poly & 0.01250 & 0.01468 & 8.10 & 8433665 \\ 
            & HDANN3-&\hspace{-1.54em}small   & 5 & 4 & 3 & ReLU & poly & 0.02018 & 0.07378 & 36.35 & 169 \\ \hline 
            & DNN&\hspace{-1.54em}-best             & 14 & 128 & & tanh & & 0.02070 & 0.02301 & 17.24 & 215681 \\ 
            & DANN&\hspace{-1.54em}-best            & 5 & 64 & 5 & & poly & 0.01538 & 0.01819 & 16.72 & 84481 \\ 
            & DANN&\hspace{-1.54em}-small     & 3 & 64 & 5 & & poly & 0.01813 & 0.02135 & 13.46 & 43393 \\ 
            & HDANN1&\hspace{-1.54em}-best          & 3 & 1024 & 3 & ReLU & poly & 0.01241 & 0.01321 & 4.77 & 2119681 \\ 
{2nd}  & HDANN1&\hspace{-1.54em}-small   & 3 & 64 & 3 & ReLU & poly & 0.01833 & 0.02230 & 10.34 & 9601 \\ 
            & HDANN2&\hspace{-1.54em}-best          & 7 & 256 & 3 & tanh & poly & 0.01590 & 0.01649 & 18.45 & 397313 \\ 
            & HDANN2&\hspace{-1.54em}-small   & 3 & 256 & 9 & ReLU & poly & 0.02059 & 0.02855 & 12.28 & 135681 \\ 
            & HDANN3&\hspace{-1.54em}-best          & 9 & 1024 & 11 & ReLU & cos & 0.01355 & 0.01352 & 25.08 & 8476673 \\ 
            & HDANN3&\hspace{-1.54em}-small   & 5 & 16 & 7 & ReLU & poly & 0.01978 & 0.02478 & 20.49 & 1889 \\ \hline 
            & DNN&\hspace{-1.54em}-best             & 10 & 128 & & tanh & & 0.02610 & 0.02591 & 15.79 & 149633 \\ 
            & DANN&\hspace{-1.54em}-best            & 5 & 1024 & 5 & & poly & 0.01456 & 0.01319 & 17.99 & 21012481 \\ 
            & DANN&\hspace{-1.54em}-small     & 5 & 4 & 7 & & poly & 0.02309 & 0.01890 & 74.93 & 665 \\ 
            & HDANN1&\hspace{-1.54em}-best          & 7 & 1024 & 7 & ReLU & poly & 0.01187 & 0.01130 & 7.75 & 6342657 \\ 
{3rd}  & HDANN1&\hspace{-1.54em}-small   & 3 & 16 & 7 & ReLU & poly & 0.02364 & 0.02176 & 12.15 & 1249 \\ 
            & HDANN2&\hspace{-1.54em}-best          & 9 & 256 & 11 & tanh & poly & 0.01865 & 0.01667 & 16.91 & 530945 \\ 
            & HDANN2&\hspace{-1.54em}-small   & 5 & 64 & 9 & tanh & poly & 0.02434 & 0.02193 & 23.82 & 17665 \\ 
            & HDANN3&\hspace{-1.54em}-best          & 9 & 1024 & 5 & tanh & poly & 0.01469 & 0.01302 & 11.89 & 8433665 \\ 
            & HDANN3&\hspace{-1.54em}-small   & 3 & 16 & 7 & tanh & poly & 0.02538 & 0.02414 & 16.85 & 1345 \\ \hline 
            & DNN&\hspace{-1.54em}-best             & 6 & 128 & & tanh & & 0.02436 & 0.02902 & 18.31 & 83585 \\ 
            & DANN&\hspace{-1.54em}-best            & 3 & 256 & 11 & & poly & 0.01415 & 0.01701 & 17.04 & 1462273 \\ 
            & DANN&\hspace{-1.54em}-small     & 3 & 16 & 5 & & poly & 0.02229 & 0.03963 & 21.83 & 3169 \\ 
            & HDANN1&\hspace{-1.54em}-best          & 5 & 1024 & 5 & ReLU & poly & 0.01086 & 0.01405 & 6.45 & 4231169 \\ 
{4th}  & HDANN1&\hspace{-1.54em}-small   & 5 & 16 & 5 & ReLU & poly & 0.02143 & 0.02873 & 16.33 & 1601 \\ 
            & HDANN2&\hspace{-1.54em}-best          & 7 & 256 & 3 & tanh & poly & 0.01710 & 0.02494 & 18.03 & 397313 \\ 
            & HDANN2&\hspace{-1.54em}-small   & 5 & 64 & 3 & tanh & poly & 0.02356 & 0.03438 & 37.03 & 17281 \\ 
            & HDANN3&\hspace{-1.54em}-best          & 9 & 64 & 5 & ReLU & poly & 0.01241 & 0.02034 & 14.32 & 35585 \\ 
            & HDANN3&\hspace{-1.54em}-small   & 7 & 4 & 9 & ReLU & poly & 0.01990 & 0.03804 & 48.33 & 377 \\ \hline 
            & DNN&\hspace{-1.54em}-best             & 8 & 128 & & tanh & & 0.02241 & 0.02759 & 19.14 & 116609 \\ 
            & DANN&\hspace{-1.54em}-best            & 5 & 1024 & 5 & & poly & 0.01560 & 0.01401 & 13.94 & 21012481 \\ 
            & DANN&\hspace{-1.54em}-small     & 5 & 16 & 9 & & poly & 0.01897 & 0.02040 & 21.15 & 10305 \\ 
            & HDANN1&\hspace{-1.54em}-best          & 5 & 1024 & 5 & ReLU & poly & 0.01295 & 0.01246 & 3.77 & 4231169 \\ 
{5th}  & HDANN1&\hspace{-1.54em}-small   & 3 & 64 & 5 & ReLU & poly & 0.01943 & 0.01983 & 6.70 & 10369 \\ 
            & HDANN2&\hspace{-1.54em}-best          & 7 & 256 & 11 & tanh & poly & 0.01688 & 0.02023 & 15.58 & 399361 \\ 
            & HDANN2&\hspace{-1.54em}-small   & 7 & 64 & 3 & tanh & poly & 0.02094 & 0.02423 & 24.07 & 25601 \\ 
            & HDANN3&\hspace{-1.54em}-best          & 9 & 1024 & 9 & ReLU & cos & 0.01385 & 0.01434 & 11.32 & 8462337 \\ 
            & HDANN3&\hspace{-1.54em}-small   & 5 & 16 & 11 & tanh & poly & 0.02213 & 0.01978 & 18.55 & 2337 \\ \hline 
\end{tabular}
\end{table}

\begin{table}[!h]
\caption{Model 2 in scenario 1}
\centering
\label{tab:Square_small_monte_top_5}
\tiny
\begin{tabular}{c|rl|ccccc|cccc}
\hline
Monte-Carlo sample & \multicolumn{2}{c|}{Network} & $L$ & $p$ & $q$ & $\sigma$ & $B_r$ & Validation Error & Test Error & Training Time (sec) & \# of Parameters \\ \hline
            & DNN&\hspace{-1.54em}-best             & 10 & 32 & & tanh & & 0.35643 & 0.26051 & 13.14 & 9761 \\ 
            & DANN&\hspace{-1.54em}-best            & 3 & 256 & 7 & & poly & 0.12137 & 0.09770 & 7.58 & 930817 \\ 
            & DANN&\hspace{-1.54em}-small     & 1 & 256 & 7 & & poly & 0.34713 & 0.35049 & 10.34 & 12801 \\ 
            & HDANN1&\hspace{-1.54em}-best          & 9 & 256 & 3 & ReLU & poly & 0.06505 & 0.07388 & 5.51 & 531457 \\ 
{1st}  & HDANN1&\hspace{-1.54em}-small   & 5 & 16 & 5 & ReLU & poly & 0.35420 & 0.24014 & 9.36 & 1601 \\ 
            & HDANN2&\hspace{-1.54em}-best          & 3 & 1024 & 9 & tanh & poly & 0.14653 & 0.12778 & 10.39 & 2115585 \\ 
            & HDANN2&\hspace{-1.54em}-small   & 5 & 64 & 5 & tanh & cos & 0.29902 & 0.18741 & 10.02 & 17409 \\ 
            & HDANN3&\hspace{-1.54em}-best          & 7 & 1024 & 7 & tanh & cos & 0.12975 & 0.09299 & 10.43 & 6348801 \\ 
            & HDANN3&\hspace{-1.54em}-small   & 7 & 16 & 11 & ReLU & poly & 0.29647 & 0.27922 & 18.93 & 2881 \\ \hline 
            & DNN&\hspace{-1.54em}-best             & 10 & 32 & & tanh & & 0.25860 & 0.32917 & 20.07 & 9761 \\ 
            & DANN&\hspace{-1.54em}-best            & 3 & 256 & 7 & & poly & 0.13784 & 0.16340 & 12.17 & 930817 \\ 
            & DANN&\hspace{-1.54em}-small     & 3 & 16 & 5 & & poly & 0.24084 & 0.29161 & 24.84 & 3169 \\ 
            & HDANN1&\hspace{-1.54em}-best          & 5 & 1024 & 3 & ReLU & poly & 0.04920 & 0.05600 & 5.11 & 4218881 \\ 
{2nd}  & HDANN1&\hspace{-1.54em}-small   & 7 & 16 & 3 & ReLU & poly & 0.22010 & 0.27233 & 15.97 & 1953 \\ 
            & HDANN2&\hspace{-1.54em}-best          & 5 & 1024 & 11 & ReLU & poly & 0.08971 & 0.14992 & 11.25 & 4216833 \\ 
            & HDANN2&\hspace{-1.54em}-small   & 3 & 64 & 7 & ReLU & cos & 0.24942 & 0.35515 & 13.10 & 9217 \\ 
            & HDANN3&\hspace{-1.54em}-best          & 7 & 256 & 7 & tanh & cos & 0.09411 & 0.15048 & 8.08 & 407553 \\ 
            & HDANN3&\hspace{-1.54em}-small   & 5 & 16 & 5 & ReLU & poly & 0.25859 & 0.35748 & 21.59 & 1665 \\ \hline 
            & DNN&\hspace{-1.54em}-best             & 8 & 128 & & tanh & & 0.29479 & 0.31256 & 13.11 & 116609 \\ 
            & DANN&\hspace{-1.54em}-best            & 3 & 256 & 7 & & poly & 0.09222 & 0.12092 & 10.83 & 930817 \\ 
            & DANN&\hspace{-1.54em}-small     & 3 & 64 & 5 & & poly & 0.15109 & 0.17232 & 14.12 & 43393 \\ 
            & HDANN1&\hspace{-1.54em}-best          & 9 & 1024 & 7 & ReLU & poly & 0.06996 & 0.08331 & 7.85 & 8441857 \\ 
{3rd}  & HDANN1&\hspace{-1.54em}-small   & 5 & 16 & 7 & ReLU & poly & 0.24158 & 0.23220 & 14.32 & 1793 \\ 
            & HDANN2&\hspace{-1.54em}-best          & 3 & 1024 & 7 & tanh & cos & 0.15238 & 0.17052 & 10.33 & 2113537 \\ 
            & HDANN2&\hspace{-1.54em}-small   & 7 & 64 & 7 & tanh & poly & 0.23311 & 0.29389 & 17.37 & 25857 \\ 
            & HDANN3&\hspace{-1.54em}-best          & 9 & 64 & 11 & ReLU & poly & 0.10450 & 0.15059 & 19.83 & 38273 \\ 
            & HDANN3&\hspace{-1.54em}-small   & 7 & 16 & 9 & tanh & poly & 0.28260 & 0.26960 & 27.38 & 2657 \\ \hline 
            & DNN&\hspace{-1.54em}-best             & 18 & 32 & & tanh & & 0.23065 & 0.23792 & 24.79 & 18209 \\ 
            & DANN&\hspace{-1.54em}-best            & 3 & 64 & 11 & & poly & 0.11417 & 0.09610 & 20.11 & 95233 \\ 
            & DANN&\hspace{-1.54em}-small     & 3 & 16 & 7 & & poly & 0.22674 & 0.17300 & 27.27 & 4417 \\ 
            & HDANN1&\hspace{-1.54em}-best          & 5 & 1024 & 3 & ReLU & poly & 0.04283 & 0.04475 & 5.51 & 4218881 \\ 
{4th}  & HDANN1&\hspace{-1.54em}-small   & 5 & 16 & 5 & ReLU & poly & 0.21227 & 0.26728 & 13.96 & 1601 \\ 
            & HDANN2&\hspace{-1.54em}-best          & 7 & 256 & 7 & tanh & cos & 0.12888 & 0.13913 & 9.53 & 398337 \\ 
            & HDANN2&\hspace{-1.54em}-small   & 5 & 64 & 7 & tanh & cos & 0.18766 & 0.23790 & 10.56 & 17537 \\ 
            & HDANN3&\hspace{-1.54em}-best          & 7 & 64 & 5 & tanh & poly & 0.09150 & 0.08640 & 15.00 & 27265 \\ 
            & HDANN3&\hspace{-1.54em}-small   & 5 & 16 & 7 & tanh & poly & 0.19589 & 0.15722 & 26.45 & 1889 \\ \hline 
            & DNN&\hspace{-1.54em}-best             & 10 & 32 & & tanh & & 0.42048 & 0.33301 & 17.30 & 9761 \\ 
            & DANN&\hspace{-1.54em}-best            & 3 & 256 & 9 & & poly & 0.13036 & 0.12474 & 10.02 & 1196545 \\ 
            & DANN&\hspace{-1.54em}-small     & 9 & 4 & 11 & & poly & 0.34201 & 0.47686 & 71.27 & 1753 \\ 
            & HDANN1&\hspace{-1.54em}-best          & 9 & 1024 & 7 & ReLU & poly & 0.05554 & 0.05279 & 10.13 & 8441857 \\ 
{5th}  & HDANN1&\hspace{-1.54em}-small   & 3 & 16 & 7 & ReLU & poly & 0.32926 & 0.38209 & 13.85 & 1249 \\ 
            & HDANN2&\hspace{-1.54em}-best          & 5 & 1024 & 3 & tanh & cos & 0.13000 & 0.12084 & 12.00 & 4208641 \\ 
            & HDANN2&\hspace{-1.54em}-small   & 9 & 16 & 11 & ReLU & cos & 0.40186 & 0.42147 & 20.92 & 2465 \\ 
            & HDANN3&\hspace{-1.54em}-best          & 9 & 1024 & 7 & ReLU & cos & 0.10662 & 0.08755 & 10.71 & 8448001 \\ 
            & HDANN3&\hspace{-1.54em}-small   & 3 & 4 & 11 & ReLU & cos & 0.41798 & 0.37439 & 34.46 & 353 \\ \hline 

\end{tabular}
\end{table}

\begin{table}[!h]
\caption{Model 2 in scenario 2}
\centering
\label{tab:Square_big_monte_top_5}
\tiny
\begin{tabular}{c|rl|ccccc|cccc}
\hline
Monte-Carlo sample & \multicolumn{2}{c|}{Network} & $L$ & $p$ & $q$ & $\sigma$ & $B_r$ & Validation Error & Test Error & Training Time (sec) & \# of Parameters \\ \hline
            & DNN&\hspace{-1.54em}-best             & 14 & 32 & & tanh & & 0.17844 & 0.23466 & 33.25 & 13985 \\ 
            & DANN&\hspace{-1.54em}-best            & 5 & 64 & 5 & & poly & 0.07885 & 0.09495 & 16.08 & 84481 \\ 
            & DANN&\hspace{-1.54em}-small     & 3 & 16 & 5 & & poly & 0.12545 & 0.18081 & 18.79 & 3169 \\ 
            & HDANN1&\hspace{-1.54em}-best          & 7 & 1024 & 5 & ReLU & poly & 0.03153 & 0.02817 & 5.26 & 6330369 \\ 
{1st}   & HDANN1&\hspace{-1.54em}-small   & 5 & 16 & 3 & ReLU & poly & 0.16609 & 0.22101 & 10.49 & 1409 \\ 
            & HDANN2&\hspace{-1.54em}-best          & 7 & 1024 & 3 & ReLU & poly & 0.05802 & 0.05988 & 7.84 & 6307841 \\ 
            & HDANN2&\hspace{-1.54em}-small   & 9 & 16 & 3 & tanh & poly & 0.17822 & 0.23816 & 38.32 & 2337 \\ 
            & HDANN3&\hspace{-1.54em}-best          & 7 & 1024 & 11 & ReLU & cos & 0.04894 & 0.06729 & 12.37 & 6377473 \\ 
            & HDANN3&\hspace{-1.54em}-small   & 5 & 16 & 3 & ReLU & poly & 0.17144 & 0.23110 & 18.03 & 1441 \\ \hline 
            & DNN&\hspace{-1.54em}-best             & 10 & 128 & & tanh & & 0.01934 & 0.02213 & 15.94 & 149633 \\ 
            & DANN&\hspace{-1.54em}-best            & 5 & 64 & 5 & & poly & 0.01405 & 0.01593 & 19.47 & 84481 \\ 
            & DANN&\hspace{-1.54em}-small     & 5 & 16 & 5 & & poly & 0.01778 & 0.02280 & 32.55 & 5761 \\ 
            & HDANN1&\hspace{-1.54em}-best          & 5 & 1024 & 7 & ReLU & poly & 0.01231 & 0.01224 & 5.04 & 4243457 \\ 
{2nd}  & HDANN1&\hspace{-1.54em}-small   & 3 & 64 & 3 & ReLU & poly & 0.01657 & 0.02008 & 8.75 & 9601 \\ 
            & HDANN2&\hspace{-1.54em}-best          & 7 & 256 & 7 & tanh & poly & 0.01630 & 0.01592 & 17.84 & 398337 \\ 
            & HDANN2&\hspace{-1.54em}-small   & 5 & 256 & 3 & ReLU & poly & 0.01835 & 0.02421 & 11.27 & 265729 \\ 
            & HDANN3&\hspace{-1.54em}-best          & 9 & 64 & 3 & tanh & poly & 0.01351 & 0.01627 & 16.02 & 34689 \\ 
            & HDANN3&\hspace{-1.54em}-small   & 7 & 16 & 5 & ReLU & poly & 0.01933 & 0.02466 & 23.30 & 2209 \\ \hline 
            & DNN&\hspace{-1.54em}-best             & 14 & 128 & & tanh & & 0.02322 & 0.02022 & 20.64 & 215681 \\ 
            & DANN&\hspace{-1.54em}-best            & 3 & 64 & 11 & & poly & 0.01450 & 0.01561 & 14.51 & 95233 \\ 
            & DANN&\hspace{-1.54em}-small     & 3 & 16 & 5 & & poly & 0.02280 & 0.01997 & 18.80 & 3169 \\ 
            & HDANN1&\hspace{-1.54em}-best          & 7 & 1024 & 7 & ReLU & poly & 0.01202 & 0.01145 & 6.52 & 6342657 \\ 
{3rd}  & HDANN1&\hspace{-1.54em}-small   & 7 & 16 & 5 & ReLU & poly & 0.02192 & 0.02014 & 13.64 & 2145 \\ 
            & HDANN2&\hspace{-1.54em}-best          & 7 & 256 & 3 & tanh & poly & 0.01808 & 0.01664 & 17.12 & 397313 \\ 
            & HDANN2&\hspace{-1.54em}-small   & 5 & 256 & 5 & tanh & poly & 0.02008 & 0.02010 & 18.50 & 266241 \\ 
            & HDANN3&\hspace{-1.54em}-best          & 9 & 256 & 5 & ReLU & poly & 0.01386 & 0.01253 & 9.38 & 535553 \\ 
            & HDANN3&\hspace{-1.54em}-small   & 5 & 16 & 7 & ReLU & poly & 0.02205 & 0.02160 & 17.55 & 1889 \\ \hline 
            & DNN&\hspace{-1.54em}-best             & 12 & 128 & & tanh & & 0.02621 & 0.03349 & 16.05 & 182657 \\ 
            & DANN&\hspace{-1.54em}-best            & 3 & 256 & 5 & & poly & 0.01407 & 0.01893 & 9.92 & 665089 \\ 
            & DANN&\hspace{-1.54em}-small     & 5 & 4 & 5 & & poly & 0.02538 & 0.04560 & 66.62 & 481 \\ 
            & HDANN1&\hspace{-1.54em}-best          & 3 & 1024 & 7 & ReLU & poly & 0.01092 & 0.01428 & 3.66 & 2144257 \\ 
{4th}  & HDANN1&\hspace{-1.54em}-small   & 3 & 16 & 5 & ReLU & poly & 0.02571 & 0.03461 & 15.51 & 1057 \\ 
            & HDANN2&\hspace{-1.54em}-best          & 9 & 256 & 11 & ReLU & poly & 0.01743 & 0.02459 & 11.17 & 530945 \\ 
            & HDANN2&\hspace{-1.54em}-small   & 5 & 64 & 3 & tanh & poly & 0.02557 & 0.03313 & 28.10 & 17281 \\ 
            & HDANN3&\hspace{-1.54em}-best          & 7 & 64 & 5 & ReLU & poly & 0.01292 & 0.02142 & 12.78 & 27265 \\ 
            & HDANN3&\hspace{-1.54em}-small   & 7 & 4 & 9 & ReLU & poly & 0.02071 & 0.04186 & 40.39 & 377 \\ \hline 
            & DNN&\hspace{-1.54em}-best             & 18 & 128 & & tanh & & 0.01702 & 0.02016 & 20.00 & 281729 \\ 
            & DANN&\hspace{-1.54em}-best            & 5 & 1024 & 5 & & poly & 0.01498 & 0.01358 & 14.66 & 21012481 \\ 
            & DANN&\hspace{-1.54em}-small     & 3 & 1024 & 5 & & poly & 0.01686 & 0.01564 & 13.91 & 10524673 \\ 
            & HDANN1&\hspace{-1.54em}-best          & 9 & 1024 & 5 & tanh & poly & 0.01326 & 0.01378 & 9.75 & 8429569 \\ 
{5th}  & HDANN1&\hspace{-1.54em}-small   & 7 & 64 & 3 & tanh & poly & 0.01428 & 0.01754 & 10.75 & 26241 \\ 
            & HDANN2&\hspace{-1.54em}-best          & 5 & 1024 & 7 & ReLU & poly & 0.01817 & 0.02410 & 10.61 & 4212737 \\ 
            & HDANN2&\hspace{-1.54em}-small   & NA & NA & NA & NA & NA & NA & NA & NA & NA \\ 
            & HDANN3&\hspace{-1.54em}-best          & 9 & 1024 & 11 & ReLU & cos & 0.01453 & 0.01370 & 14.04 & 8476673 \\ 
            & HDANN3&\hspace{-1.54em}-small   & 7 & 64 & 5 & ReLU & poly & 0.01606 & 0.01640 & 10.73 & 27265 \\ \hline 
\end{tabular}
\end{table}

\clearpage

\subsection{Selected hyperparameters for Table \ref{tab:data analysis_result}}\label{appB}

\begin{table}[!h]
\centering
\label{tab:California_top}
\tiny
\begin{tabular}{c|rl|ccccc|cccc}
\hline
Fold($r$) & \multicolumn{2}{c|}{Network} & $L$ & $p$ & $q$ & $\sigma$ & $B_r$ & Validation Error & Test Error & Training Time (sec) & \# of Parameters \\ \hline
            & DNN&\hspace{-1.54em}-best             & 8 & 2048 & & tanh & & 0.36354 & 0.36452 & 29.77 & 29394945 \\
            & DANN&\hspace{-1.54em}-best            & 3 & 256 & 3 & & cos & 0.28369 & 0.29679 & 24.64 & 400897 \\
            & DANN&\hspace{-1.54em}-small           & 1 & 16 & 9 & & cos & 0.34448 & 0.34292 & 47.59 & 1313 \\
            & HDANN1&\hspace{-1.54em}-best          & 5 & 256 & 9 & tanh & cos & 0.23888 & 0.25104 & 31.65 & 282113 \\
{1st}       & HDANN1&\hspace{-1.54em}-small         & 1 & 16 & 11 & tanh & cos & 0.35796 & 0.35257 & 38.75 & 1441 \\
            & HDANN2&\hspace{-1.54em}-best          & 7 & 256 & 7 & ReLU & cos & 0.26842 & 0.26609 & 29.61 & 398849 \\
            & HDANN2&\hspace{-1.54em}-small         & 5 & 16 & 9 & ReLU & cos & 0.36243 & 0.34882 & 47.86 & 1377 \\
            & HDANN3&\hspace{-1.54em}-best          & 9 & 256 & 11 & tanh & cos & 0.26357 & 0.27910 & 42.70 & 551937 \\
            & HDANN3&\hspace{-1.54em}-small         & 1 & 16 & 11 & ReLU & cos & 0.35315 & 0.34964 & 37.90 & 1873 \\ \hline 
            & DNN&\hspace{-1.54em}-best             & 10 & 512 & & tanh & & 0.35576 & 0.36719 & 35.76 & 2369025 \\
            & DANN&\hspace{-1.54em}-best            & 3 & 256 & 3 & & cos & 0.28919 & 0.30267 & 35.17 & 400897 \\
            & DANN&\hspace{-1.54em}-small           & 1 & 16 & 9 & & cos & 0.34481 & 0.35719 & 42.74 & 1313 \\
            & HDANN1&\hspace{-1.54em}-best          & 5 & 256 & 5 & tanh & cos & 0.24301 & 0.25160 & 45.46 & 273921 \\
{2nd}       & HDANN1&\hspace{-1.54em}-small         & 7 & 4 & 11 & ReLU & cos & 0.34772 & 0.35618 & 91.18 & 481 \\
            & HDANN2&\hspace{-1.54em}-best          & 5 & 1024 & 3 & ReLU & cos & 0.26089 & 0.27114 & 26.28 & 4210689 \\
            & HDANN2&\hspace{-1.54em}-small         & 5 & 16 & 7 & ReLU & cos & 0.33417 & 0.35107 & 91.46 & 1345 \\
            & HDANN3&\hspace{-1.54em}-best          & 7 & 256 & 9 & tanh & cos & 0.27156 & 0.28325 & 43.75 & 415745 \\
            & HDANN3&\hspace{-1.54em}-small         & 1 & 16 & 9 & ReLU & cos & 0.35511 & 0.37081 & 55.81 & 1585 \\ \hline 
            & DNN&\hspace{-1.54em}-best             & 16 & 512 & & tanh & & 0.39892 & 0.41335 & 42.62 & 3944961 \\
            & DANN&\hspace{-1.54em}-best            & 5 & 256 & 3 & & cos & 0.29730 & 0.30946 & 36.29 & 794625 \\
            & DANN&\hspace{-1.54em}-small           & 1 & 4 & 9 & & cos & 0.39005 & 0.40920 & 57.34 & 329 \\
            & HDANN1&\hspace{-1.54em}-best          & 7 & 256 & 7 & tanh & cos & 0.25106 & 0.26462 & 40.85 & 409601 \\
{3rd}       & HDANN1&\hspace{-1.54em}-small         & 3 & 4 & 9 & tanh & cos & 0.39860 & 0.41475 & 52.15 & 337 \\
            & HDANN2&\hspace{-1.54em}-best          & 7 & 1024 & 3 & ReLU & poly & 0.27795 & 0.29895 & 38.38 & 6309889 \\
            & HDANN2&\hspace{-1.54em}-small         & 7 & 16 & 7 & ReLU & cos & 0.38553 & 0.39270 & 72.36 & 1889 \\
            & HDANN3&\hspace{-1.54em}-best          & 9 & 256 & 7 & tanh & cos & 0.27094 & 0.28048 & 52.14 & 542721 \\
            & HDANN3&\hspace{-1.54em}-small         & 5 & 4 & 9 & tanh & cos & 0.38966 & 0.40568 & 60.28 & 409 \\ \hline 
            & DNN&\hspace{-1.54em}-best             & 14 & 512 & & tanh & & 0.36918 & 0.35756 & 49.35 & 3419649 \\
            & DANN&\hspace{-1.54em}-best            & 5 & 64 & 3 & & cos & 0.33428 & 0.32911 & 31.39 & 51201 \\
            & DANN&\hspace{-1.54em}-small           & 1 & 16 & 9 & & cos & 0.35228 & 0.33861 & 57.29 & 1313 \\
            & HDANN1&\hspace{-1.54em}-best          & 5 & 256 & 7 & tanh & cos & 0.26450 & 0.25164 & 49.93 & 278017 \\
{4th}       & HDANN1&\hspace{-1.54em}-small         & 1 & 16 & 11 & ReLU & cos & 0.35975 & 0.34238 & 49.34 & 1441 \\
            & HDANN2&\hspace{-1.54em}-best          & 9 & 1024 & 3 & ReLU & poly & 0.30501 & 0.29238 & 31.69 & 8409089 \\
            & HDANN2&\hspace{-1.54em}-small         & 5 & 64 & 9 & ReLU & cos & 0.36121 & 0.35440 & 65.02 & 17793 \\
            & HDANN3&\hspace{-1.54em}-best          & 9 & 256 & 9 & ReLU & cos & 0.29437 & 0.28692 & 29.48 & 547329 \\
            & HDANN3&\hspace{-1.54em}-small         & 1 & 16 & 11 & ReLU & cos & 0.36017 & 0.34910 & 52.47 & 1873 \\ \hline 
            & DNN&\hspace{-1.54em}-best             & 8 & 512 & & tanh & & 0.42504 & 0.41224 & 25.95 & 1843713 \\
            & DANN&\hspace{-1.54em}-best            & 3 & 64 & 3 & & cos & 0.31279 & 0.31549 & 46.84 & 26497 \\
            & DANN&\hspace{-1.54em}-small           & 1 & 4 & 7 & & cos & 0.39868 & 0.40350 & 41.10 & 257 \\
            & HDANN1&\hspace{-1.54em}-best          & 7 & 256 & 11 & tanh & cos & 0.23787 & 0.24019 & 57.01 & 417793 \\
{5th}       & HDANN1&\hspace{-1.54em}-small         & 3 & 4 & 7 & ReLU & cos & 0.41564 & 0.41419 & 55.44 & 273 \\
            & HDANN2&\hspace{-1.54em}-best          & 9 & 1024 & 11 & ReLU & poly & 0.30263 & 0.31050 & 46.30 & 8417281 \\
            & HDANN2&\hspace{-1.54em}-small         & 5 & 16 & 7 & ReLU & cos & 0.42393 & 0.42090 & 44.43 & 1345 \\
            & HDANN3&\hspace{-1.54em}-best          & 7 & 256 & 7 & tanh & cos & 0.26936 & 0.27074 & 41.00 & 411137 \\
            & HDANN3&\hspace{-1.54em}-small         & 1 & 4 & 9 & ReLU & cos & 0.41760 & 0.42425 & 51.06 & 349 \\ \hline
\end{tabular}
\end{table}


\clearpage

\begin{thebibliography}{9}
\bibitem[Bauer and Kohler(2019)]{Bauer and Kohler (2019)} Bauer, B. and Kohler, M. (2019). On deep learning as a remedy for the curse of dimensionality in nonparametric regression. {\it Annals of Statistics}, \textbf{47}, 2261--2285.
\bibitem[Bilodeau(1992)]{Bilodeau (1992)} Bilodeau, M. (1992). Fourier smoother and additive models. {\it Canadian Journal of Statistics}, \textbf{20}, 241--351.
\bibitem[Cybenko(1989)]{Cybenko (1989)} Cybenko, G. (1989). Approximation by superpositions of a sigmoidal function. {\it Mathematics of Control, Signals, and Systems}, \textbf{2}, 303--314.
\bibitem[Fakhoury et al.(2022)]{Fakhoury et al. (2022)} Fakhoury, D., Fakhoury, E. and Speleers, H. (2022). ExSpliNet: An interpretable and expressive spline-based neural network. {\it Neural Networks}, \textbf{152}, 332--346.
\bibitem[Glorot and Bengio(2010)]{Glorot and Bengio (2010)} Glorot, X. and Bengio, Y. (2010). Understanding the difficulty of training deep feedforward neural networks. {\it AISTATS}, \textbf{9}, 249--256.
\bibitem[Haar(1910)]{Haar (1910)} Haar, A. (1910). Zur Theorie der orthogonalen Funktionensysteme. {\it Mathematische Annalen}, \textbf{69}, 331--371.
\bibitem[Horowitz and Mammen(2007)]{Horowitz and Mammen (2007)} Horowitz, J. L. and Mammen, E. (2007). Rate-optimal estimation for a general class of nonparametric regression models with unknown link functions. {\it Annals of Statistics}, \textbf{35}, 2589--2619.
\bibitem[Jeon and Park(2020)]{Jeon and Park (2020)} Jeon, J. M. and Park, B. U. (2020). Additive regression with Hilbertian responses. {\it Annals of Statistics}, \textbf{48}, 2671--2697.
\bibitem[Jeon et al.(2022)]{Jeon et al. (2022)} Jeon, J. M., Lee, Y. K., Mammen, E. and Park, B. U. (2022). Locally polynomial Hilbertian additive regression. {\it Bernoulli}, \textbf{28}, 2034--2066.
\bibitem[Kidger and Lyons (2020)]{Kidger and Lyons (2020)} Kidger, P. and Lyons, T. (2020). Universal Approximation with Deep Narrow Networks. {\it Proceedings of Machine Learning Research}, \textbf{125}, 1--22.
\bibitem[Kingma and Ba(2017)]{Kingma and Ba (2017)} Kingma, D. P. and Ba, J, M. (2017). Adam: A Method for Stochastic Optimization. {\it arXiv:1412.6980v9}.
\bibitem[Kohler and Langer(2021)]{Kohler and Langer (2021)} Kohler, M. and Langer, S. (2021). On the rate of convergence of fully connected deep neural network regression estimates. {\it Annals of Statistics}, \textbf{49}, 2231--2249.
\bibitem[Kolmogorov(1957)]{Kolmogorov (1957)} Kolmogorov, A. N. (1957). On the representation of continuous functions of many variables by superposition of continuous functions of one variable and addition. {\it Doklady Akademii Nauk SSSR}, \textbf{114}, 953--956.
\bibitem[LeCun et al.(2015)]{LeCun et al. (2015)} LeCun, Y., Bengio, Y. and Hinton, G. (2015). Deep learning. {\it Nature}, \textbf{521}, 436--444.
\bibitem[Linton and Nielsen(1995)]{Linton and Nielsen (1995)} Linton, O. and Nielsen, J. P. (1995). A kernel method of estimating structured nonparametric regression based on marginal integration. {\it Biometrika}, \textbf{82}, 93--101.
\bibitem[Liu et al.(2024)]{Liu et al. (2024)} Liu, Z., Wang, Y., Vaidya, S., Ruehle, F., Halverson, J., Solja\v{c}i\'{c}, M., Hou, T. Y. and Tegmark, M. (2024). KAN: Kolmogorov-Arnold Networks. {\it arXiv:2404.19756v4}.
\bibitem[Mammen et al.(1999)]{Mammen et al. (1999)} Mammen, E., Linton, O. B. and Nielsen, J. P. (1999). The existence and asymptotic properties of a backfitting projection algorithm under weak conditions. {\it Annals of Statistics}, \textbf{27}, 1443--1490.
\bibitem[McLaughlin(1969)]{McLaughlin (1969)} McLaughlin, J. R. (1969). Haar Series. {\it Transactions of the American Mathematical Society}, \textbf{137}, 153--176.
\bibitem[Meier et al.(2009)]{Meier et al. (2009)} Meier, L., van de Geer, S. and B\"{u}hlmann, P. (2009). High-dimensional additive modeling. {\it Annals of Statistics}, \textbf{37}, 3779--3821. 
\bibitem[Opsomer and Ruppert(1997)]{Opsomer and Ruppert (1997)} Opsomer, J. D. and Ruppert, D. (1997). Fitting a bivariate additive model by local polynomial regression. {\it Annals of Statistics}, \textbf{25}, 186--211.
\bibitem[Pereyra and Ward(2012)]{Pereyra and Ward (2012)} Pereyra, M. C. and Ward, L. A. (2012). {\it Harmonic Analysis: From Fourier to Wavelets}. American Mathematical Society.
\bibitem[Sardy and Ma(2024)]{Sardy and Ma (2024)} Sardy, S. and Ma, X. (2024). Sparse additive models in high dimensions with wavelets. {\it Scandinavian Journal of Statistics}, \textbf{51}, 89–-108.
\bibitem[Schmidt-Hieber(2020)]{Schmidt-Hieber (2020)} Schmidt-Hieber, J. (2020). Nonparametric regression using deep neural networks with ReLU activation function. {\it Annals of Statistics}, \textbf{48}, 1875--1897.
\bibitem[Schmidhuber(2015)]{Schmidhuber (2015)} Schmidhuber, J. (2015). Deep learning in neural networks: An overview. {\it Neural Networks}, \textbf{61}, 85--117.
\end{thebibliography}
\end{document}